\documentclass[journal]{IEEEtran}

\usepackage{times}
\usepackage{epsfig}
\usepackage{graphicx}
\usepackage{amsmath}
\usepackage{amssymb}
\usepackage{multirow}

\def\ie{\emph{i.e}. }

\def\etal{\emph{et al}. }
\def\wrt{\emph{w.r.t}. }

\newtheorem{myrem}{Remark}
\newtheorem{myrec}{Recall}

\pdfoutput=1

\ifCLASSINFOpdf
\else
\fi

\begin{document}
%
\title{On the Relation between Color Image Denoising and Classification}
%
%
%

\author{Jiqing~Wu,
        Radu~Timofte,~\IEEEmembership{Member,~IEEE},
        Zhiwu~Huang,~\IEEEmembership{Member,~IEEE},
        and~Luc~Van~Gool,~\IEEEmembership{Member,~IEEE}
\thanks{J. Wu, R. Timofte, Z. Huang and L. Van Gool are with the Department of Information Technology and Electrical Engineering, ETH Zurich, Switzerland, e-mail: \{jwu, radu.timofte, zhiwu.huang, vangool\}@vision.ee.ethz.ch}}
\maketitle

\begin{abstract}
Large amount of image denoising literature focuses on single channel images and often experimentally validates the proposed methods on tens of images at most. In this paper, we investigate the interaction between denoising and classification on large scale dataset. 
Inspired by classification models, we propose a novel deep learning architecture for color (multichannel) image denoising and report on thousands of images from ImageNet dataset as well as commonly used imagery. We study the importance of (sufficient) training data, how semantic class information can be traded for improved denoising results.
As a result, our method greatly improves PSNR performance by 0.34 - 0.51 dB on average over state-of-the art methods on large scale dataset.
We conclude that it is beneficial to incorporate in classification models.
On the other hand, we also study how noise affect classification performance. 
In the end, we come to a number of interesting conclusions, some being counter-intuitive.
\end{abstract}


%
\IEEEpeerreviewmaketitle

\section{Introduction}
Image denoising is a fundamental image restoration task. Typically it is modeled as 
\begin{equation}
\label{eq:noise}
\mathbf{Y} = \mathbf{X} + \mathbf{N},
\end{equation}
where the goal is to recover the clean image $\mathbf{X}$ from the corrupted image $\mathbf{Y}$ affected by the noise $\mathbf{N}$ assumed to be white Gaussian noise. 

A large body of work 
~\cite{Dabov-TIP-2007,Mairal-ICCV-2009,Zoran-ICCV-2011,chen2013revisiting,schmidt2014cascades,Gu-CVPR-2014,schmidt2014shrinkage} has been published in this area throughout the years. 
The proposed methods involve all kinds of techniques, varying from the use of self (patch) similarities to random fields formulations. With the recent resurgence of neural networks, researchers also applied shallow and deep convolutional neural networks (CNN) to solve the denoising problem and achieved impressive results~\cite{Burger-CVPR-2012,chen2015learning,vemulapalli2015deep}. 
It is worth mentioning that most of the proposed methods in denoising focus on single channel (gray scale) images. The color image denoising is treated as an application of straightforward modified versions of the solutions developed for single channel images and more often the multiple channels are denoised separately by deploying the single channel solutions.
A series of Dabov~\etal works on denoising are a good example of methods developed for single channel denoising and further modified for handling color images. The block-matching and 3D collaborative filtering (BM3D) method for image denoising of Dabov~\etal~\cite{Dabov-TIP-2007} was further extended to Color BM3D (CBM3D) by the same authors~\cite{dabov2007color}. Another top color image denoising has been recently proposed by Rajwade~\etal~\cite{rajwade2013image}. They applied higher order singular value decomposition (HOSVD) and reached competitive results at the cost of heavy computational load. 

On the other hand, breakthrough results were achieved by the recent and rapid development of the deep CNNs~\cite{krizhevsky2012imagenet,lecun1989backpropagation} on various vision tasks such as image classification, recognition, segmentation, and scene understanding. 
Large CNN models~\cite{krizhevsky2012imagenet,szegedy2015going,simonyan2014very} with millions of parameters were proven to significantly improve the accuracy over the non-CNN previous solutions. Even deeper CNN architectures~\cite{he2015deep} have been shown to further improve the performance.

A key contributing factor to the success of CNN, besides the hardware GPU advances, is the introduction of large scale datasets such as ImageNet~\cite{deng2009imagenet}, COCO~\cite{lin2014microsoft}, and Places~\cite{zhou2014learning} for classification, detection, segmentation, and retrieval.
The availability of millions of train images supports the development of complex and sophisticated models with millions of learned parameters. At the same time the large datasets are reliable and convincing benchmarks for validating the proposed methods.
In comparison, the image denoising works still conduct their validation experiments on surprisingly small datasets. For example, in a very recent work~\cite{vemulapalli2015deep} 400 images are used for training and 300 images for testing, images selected from~\cite{everingham2015pascal,martin2001database}, while the BM3D method~\cite{Dabov-TIP-2007} was introduced on dozens of images.
In the light of the existing significant data scale gap between the low (\ie image denoising) and high level vision tasks (\ie classification, detection, retrieval) and the data driven advances in the latter tasks we can only ask ourselves: is it really sufficient to validate denoising methods on small datasets?

Commonly the researchers develop image classification algorithms based on the assumption that the images are clean and uncorrupted. However, most of the images are corrupted and contaminated by noise. The sources of corruption are diverse, they can be due to factors such as: suboptimal use of camera sensors and settings, improper environmental conditions at capturing time, post-processing, and image compression artifacts.
More importantly, there are evidences~\cite{szegedy2013intriguing} showing that CNN models are highly nonlinear and very sensitive to slight perturbation on the image. 
It leads to the phenomenon that neural network can be easily fooled~\cite{nguyen2015deep} by manually generated images.
Hence, a study of how noisy images and denoising methods impact on the CNN model performance is necessary.

In summary, our paper is an attempt to bridge image denoising and image classification, and our 
main contributions are as follows:
\begin{enumerate}
\item We propose a novel deep architecture for denoising which incorporates designs used in image classification and largely improve over state-of-the-art methods.
\item We are the first, to the best of our knowledge, to study denoising methods on a scale of million of images.
\item We conduct a thorough investigation on how Gaussian noise affects the classification models and how the semantic information can help for improving the denoising results. 
\end{enumerate}

\subsection{Related Work}
In the realm of image denoising the self-similarities found in a natural image are widely exploited by state-of-the-art methods such as block matching and 3D collaborative filtering method (BM3D) of Dabov~\etal~\cite{Dabov-TIP-2007} and its color version CBM3D~\cite{dabov2007color}. 
The main idea is to group image patches which are similar in shape and texture.
(C)BM3D collaboratively filters the patch groups by shrinkage in a 3D transform domain to 
produce a sparse representation of the true signal in the group.
Later, Rajwade~\etal~\cite{rajwade2013image} applied the same idea and grouped the similar patches from a noisy image into a 3D stack to then compute the high-order singular value decomposition (HOSVD) coefficients of this stack. At last, they inverted the HOSVD transform to obtain the clean image.
HOSVD has a high time complexity which renders the method as very slow~\cite{rajwade2013image}.

Nowadays most of the visual data is actually tensor (\ie color image and video) instead of matrix (\ie grayscale image). 
Though traditional CNN models with 2D spatial filter were considered to be sufficient and achieved good results, in certain scenarios, high dimensional/rank filter becomes necessary to extract important features from tensor.
Ji~\etal~\cite{ji20133d} introduced a CNN with 3-dimensional filter (3DCNN) and demonstrated superior performance to the traditional 2D CNN on two action recognition benchmarks. 
In their 3DCNN model, the output $g_i$ of feature map at $(x, y, z)$ position at $i$-th CNN layer is computed as follows:
\begin{equation}
\label{3D}
    g_i^{x, y, z} = \mathop{\text{tanh}}(b_i + \mathop{\sum}_{l=0}^{L-1}\mathop{\sum}_{m=0}^{M-1}\mathop{\sum}_{n=0}^{N-1} w_{i,k}^{l,m,n}g_{i-1}^{(x+l), (y+m), (z+n)}) 
\end{equation}
where the temporal and spatial size of kernel is $L$ and $M \times N $ respectively.

\section{Proposed method}
We propose a two stage architecture for color image denoising as depicted in Fig.~\ref{fig:3DCF}.
First we convolve the color image with high pass filters to capture the high frequencies and apply our residual CNN model to obtain the intermediate result.
In the second stage, we adapt either AlexNet~\cite{krizhevsky2012imagenet} or VGG-16~\cite{simonyan2014very} deep architecture from image classification and stack it on top of our first stage model, along with introducing a novel cost function inspired by Sobolev space equipped norm~\cite{brezis2010functional}.
As the experiments will show, our proposed method overcomes the regress-to-mean problem and enables the recovery of high frequency details better than other denoising works.

\begin{table}[th!]
\centering
\caption{High pass filters for image pre-processing}
\label{hp}
\resizebox{0.8\columnwidth}{!}{
\begin{tabular}{|c|c|c|c|c|c|c|c|c|c|c|c|c|c|c|}
\cline{1-3} \cline{5-7} \cline{9-11} \cline{13-15}
0 & 0.5 & 0 &  & 0 & 0  & 0 &  & 0 & 0  & 0 &  & 0   & 0  & 0   \\ \cline{1-3} \cline{5-7} \cline{9-11} \cline{13-15} 
0 & -1  & 0 &  & 0 & -1 & 0 &  & 0 & -1 & 1 &  & 0.5 & -1 & 0.5 \\ \cline{1-3} \cline{5-7} \cline{9-11} \cline{13-15} 
0 & 0.5 & 0 &  & 0 & 1  & 0 &  & 0 & 0  & 0 &  & 0   & 0  & 0   \\ \cline{1-3} \cline{5-7} \cline{9-11} \cline{13-15}
\end{tabular}
}
\end{table}

\begin{figure}[th!]
  \centering
\includegraphics[width=0.5\textwidth]{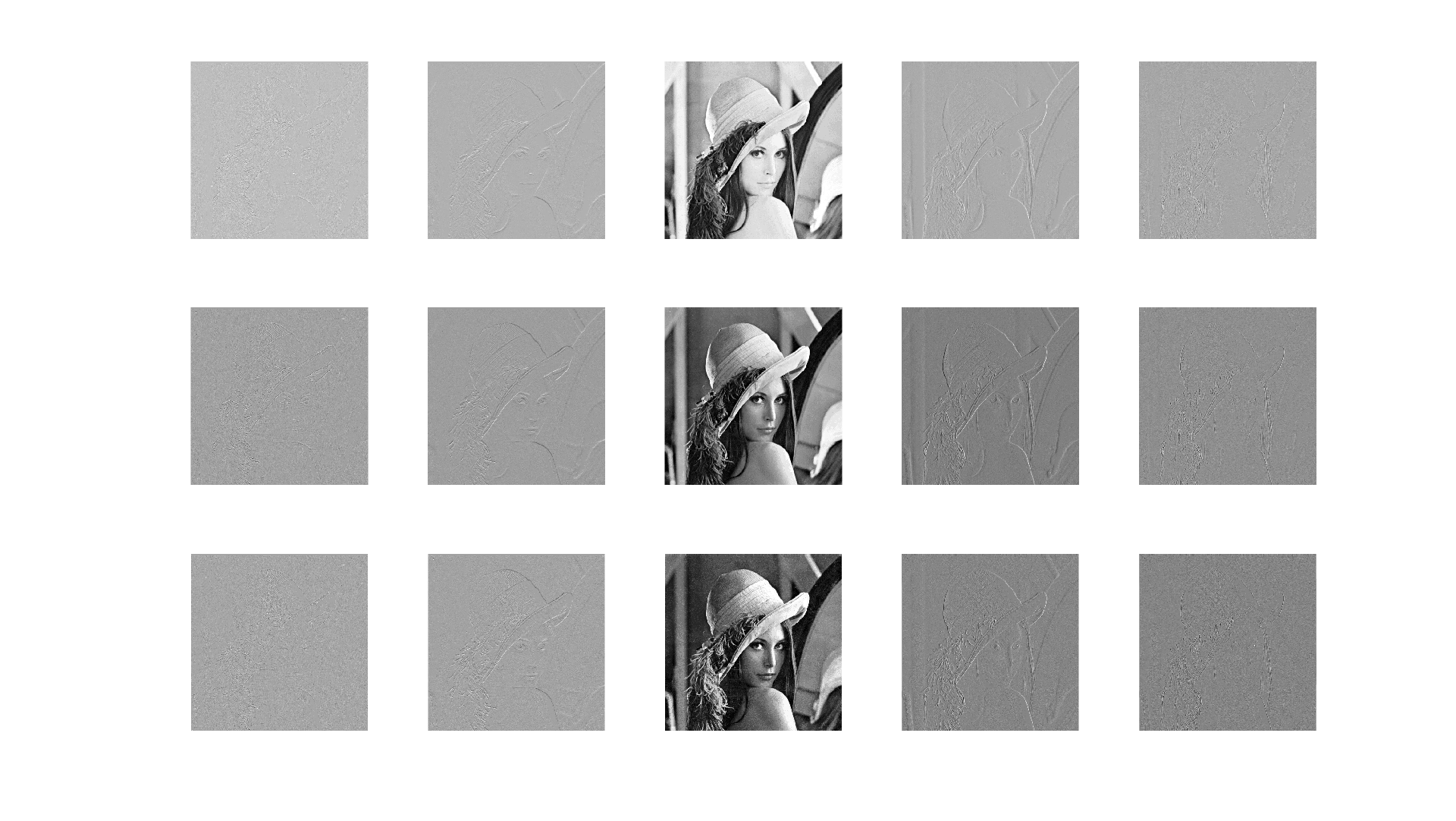}\\
\vspace{-0.4cm}
\caption{Example of pre-processed input with the high pass filters from Table~\ref{hp}}
\label{fig:gi}
\vspace{-0.5cm}
\end{figure}

\subsection{First Stage}
\textbf{Image Pre-processing} 
In the state-of-the-art (C)BM3D and HOSVD methods the group matching step is undermined by the noise. The higher the noise level is the finding of similar patches gets more difficult. Moreover, many denoising methods have the tendency to filter/regress to mean and to lose the high frequencies, \ie, the output are simply the local mean of the (highly) corrupted image.
To address the above issues, we apply the high pass filters (see Tab.~\ref{hp}) to each channel of the noisy input. Such operations correspond to the first and second directional derivative \wrt x and y direction which highlight the high frequencies.
The filtered channel responses are concatenated with the noisy channel for each of the image channels and grouped together. The assumption is that the channels of the image are highly correlated and this information can be exploited. See Fig.~\ref{fig:gi} for RGB channels and corresponding filter responses of `Lena'. 

\begin{figure*}[btp!]
    \centering
    \includegraphics[width=\textwidth,height=4cm]{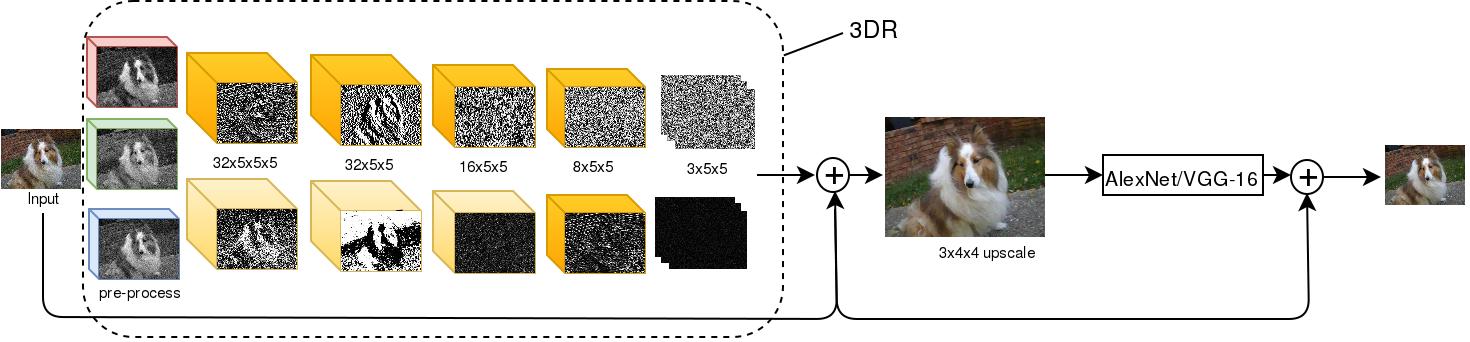}\\
    \vspace{-0.3cm}
    \caption{\textsf{Outline of proposed model.}}
    \label{fig:3DCF}
    \vspace{-0.5cm}
\end{figure*}

\textbf{Learning correlation by high rank filters} As mentioned before, color image is actually tensor data. If for grayscale image it is sufficient to apply spatial filters to extract useful features, for color image the inter-channel correlation is key to the denoising quality. Our later experiments confirms that denoising each image channel independently leads to much poorer performance.
Furthermore, after our pre-processing step we gain extra image gradients information,
which means our input currently has width, height, channel and gradient dimension.  
Thus, it does not suffice to use spatial filters in our case. 
In order to utilize the correlation among RGB channels as well as gradients, 
we are motivated to apply high rank filters to convolve high rank input.
This computation has a nice interpretation based on tensor calculus,
which we give a brief introduction here.
\begin{myrec}
A tensor $T$ of rank $(r,s)$ is a multilinear map 
\begin{equation}
  T:\mathbf{V}^* \times \mathbf{V}^* \times \cdots \times \mathbf{V}^* \times \mathbf{V} \times \mathbf{V} \times \cdots \times \mathbf{V} \to \mathbb{R} \\
\end{equation}
with $s$ copies of vector space $\mathbf{V}$ and $r$ copies of its dual $\mathbf{V}^*$.
\end{myrec}
We also consider the set of all tensors of rank $(r,s)$ is a vector space over $\mathbb{R}$, equipped with the pointwise addition and scalar multiplication. 
This vector space is denoted by $\mathbf{V} \otimes \mathbf{V} \otimes \cdots \otimes \mathbf{V} \otimes \mathbf{V}^* \otimes \mathbf{V}^* \otimes \cdots \otimes \mathbf{V}^*$ with $r$ copies of $\mathbf{V}$ and $s$ copies of $\mathbf{V}^*$.
Moreover, we use Einstein's summation convention $a_rb^r = \sum_{i=0}^n a_ib^i$.
Suppose $T$ has rank $(2, 1)$, then we define 
\begin{equation}
T^{jk}_{i} = T(\mathbf{e}_j, \mathbf{e}_k, \mathbf{e}^i),
\end{equation}
and we can express $T = T^{jk}_{i} \mathbf{e}_j \otimes \mathbf{e}_k \otimes \mathbf{e}^i$, where $\mathbf{e}_n, \mathbf{e}^n$ are the basis \wrt $\mathbf{V}, \mathbf{V}^*$.
\begin{myrem}
(Contraction) New tensors of rank $(r-1,s-1)$ can be produced by summing over one upper and low index of $T$ and obtain $T^{a_1 \cdots a_{i-1} a a_{i+1} \cdots a_r}_{b_1 \cdots b_{j-1} a b_{j+1} \cdots b_s}$.
\end{myrem}
Now the convolution of $n$-dimensional filter (rank $(0,n)$) can be considered as the contraction of tensor product between kernel $w$ and small patch $g$ sliced from the input tensor, that is,
\begin{equation}
g^{a_1 \cdots a_n} w_{b_1 \cdots b_n}\mathbf{e}_{a_1} \otimes \cdots \otimes \mathbf{e}_{a_n} \mathbf{e}^{b_1} \otimes \cdots \otimes  \mathbf{e}^{b_n} \mapsto g^{a_1 \cdots a_n} w_{a_1 \cdots a_n}.
\end{equation}

Both width and depth of the CNN matter~\cite{zagoruyko2016wide,he2015deep} and considering the trade-off between time complexity and memory usage,
our first stage (3DR) is designed for ensemble learning and has two 5-layers with same architecture (See Fig.~\ref{fig:3DCF}). 
To incorporate the preprocessed image tensors in our two CNN models, 
we use the n-dimensional filters to convolve the input, so that we can intensively explore the high frequency information which are greatly contaminated by noise. 
For efficiency we recommend 3D filters for the first layer of both two CNN models. 

We chose tanh instead of ReLU as the activation function mainly for two reasons. 
First, the negative update is necessary for computing the residual in our denoising task,
while ReLU simply ignores the negative value.
Second, tanh acts like a normalization of feature map output by bringing the value into the interval $[-1, 1]$ and the extreme outputs will not occur during the training.

Recently, learning residuals has been proved to be a simple and useful trick in previous works~\cite{he2015deep,kim2015accurate}.
Hence, instead of predicting denoised image, we estimate the image residual.
In the end, we simply average the output residuals from the two 5-layer CNN networks $\mathcal{R}_{1,2}$, then obtain the intermediate denoised image $\hat{\mathbf{X}}$ by adding averaged residuals $\mathcal{R}_{1,2}(\mathbf{Y})$ and noisy image$\mathbf{Y}$, 
\begin{equation}
\hat{\mathbf{X}} = \lambda_1\mathcal{R}_1(\mathbf{Y}) + (1 - \lambda_1) \mathcal{R}_2(\mathbf{Y}) + \mathbf{Y},
\end{equation}
where $\lambda_1$ is fixed to $0.5$ in all our experiments.
For better summarizing the property of our first stage, we call it 3D residual learning stage (3DR).

\begin{figure*}
    \centering
    \resizebox{\linewidth}{!}
    {
    \begin{tabular}{ccccc}
    \includegraphics[width=0.4\linewidth]{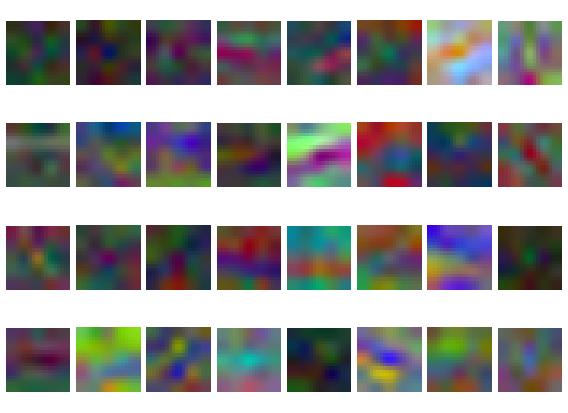}&
    \includegraphics[width=0.4\linewidth]{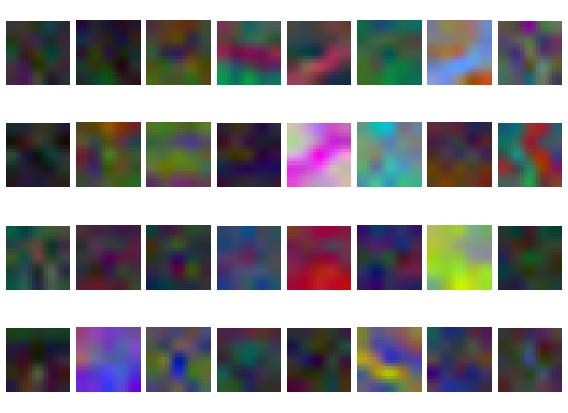}&
    \includegraphics[width=0.4\linewidth]{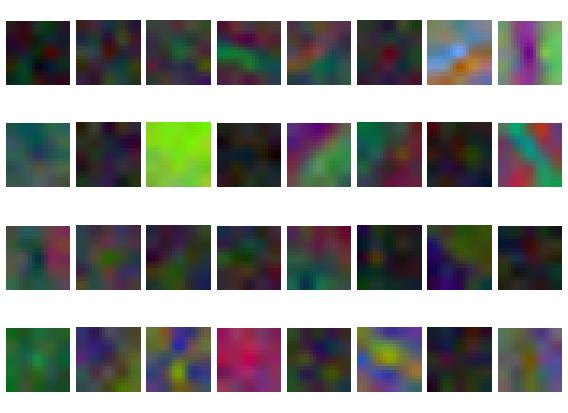}&
    \includegraphics[width=0.4\linewidth]{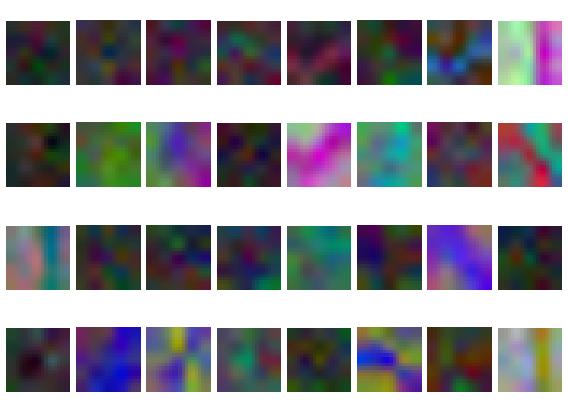}&
    \includegraphics[width=0.4\linewidth]{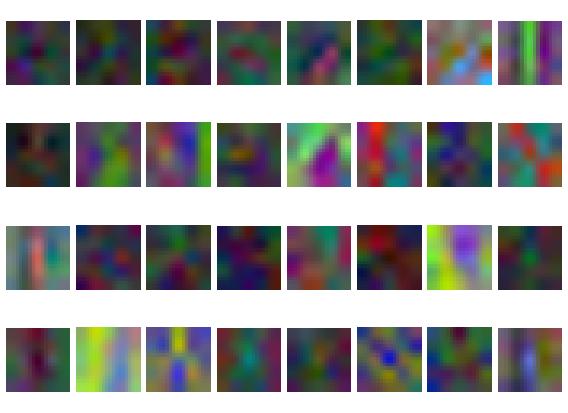}\\
    2nd derivative filter along y&
    1st derivative filter along y&
    without filter&
    1st derivative filter along x&
    2nd derivative filter along x\\
    \end{tabular}
    }
    \vspace{-0.2cm}
    \caption{Visualization of the first layer of 3DR in our model Best zoom on screen.}
    \label{fig:3dr}
    \vspace{-0.3cm}
\end{figure*}

\begin{figure*}[ht!]
    \centering
    \resizebox{0.85\linewidth}{!}
    {
    \begin{tabular}{cccc}
    \includegraphics[width=0.4\linewidth]{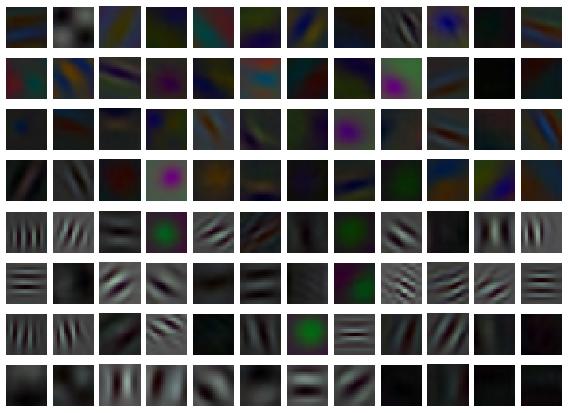}&
    \includegraphics[width=0.4\linewidth]{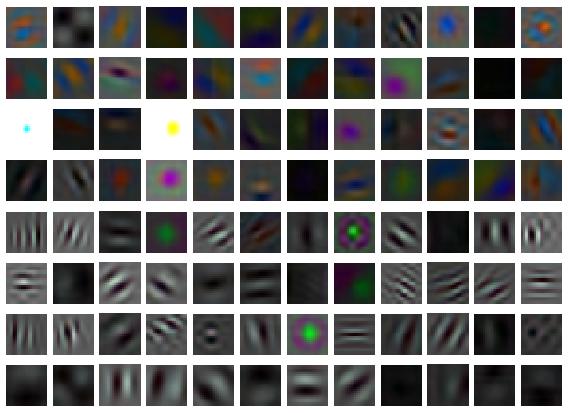}&
    \includegraphics[width=0.4\linewidth]{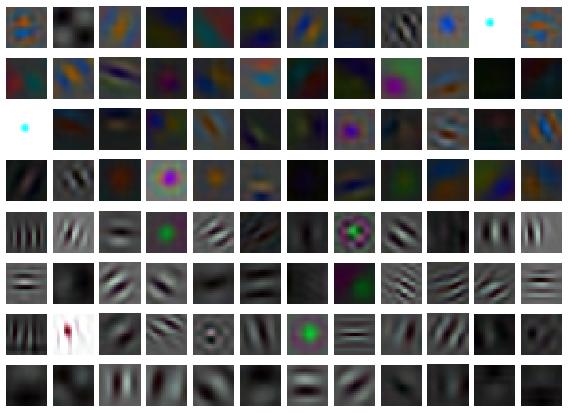}&
    \includegraphics[width=0.4\linewidth]{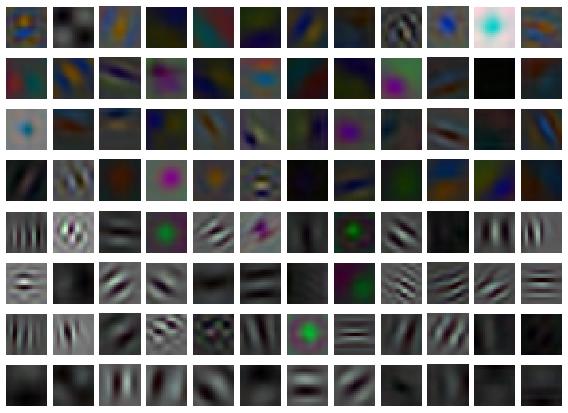}\\    
    Original filters& $\sigma = 15$ & $\sigma = 25$ & $\sigma = 50$\\
    \end{tabular}
    }
    \caption{Visualization of the first layer of AlexNet. Best zoom on screen.}
    \label{fig:alex}
    \vspace{-0.3cm}
\end{figure*}

\subsection{Second Stage}
\textbf{Collaboration with classification architecture} 
Inspired by~\cite{he2015deep}, we fully take advantage of sophisticated CNN architectures proposed for image classification and adapt them on top of our 3DR model as a second stage.
We adapt AlexNet~\cite{krizhevsky2012imagenet} and VGG-16~\cite{simonyan2014very} which are two widely used CNN architectures. 
We intend to have an end-to-end model and an output denoised image with the same size as the input image. 
In AlexNet/VGG-16 the stride set in the pooling and convolution layer causes image size reduction. Therefore, we upscale the first stage denoised image with a deconvolution layer and keep only one max pooling layer for AlexNet/VGG-16, such that the size of output image keeps the same.
Due to memory constraints, we use only a part of the VGG-16 model up to conv3-256 layer (see~\cite{simonyan2014very}).
Additionally, we replace the fully connected and softmax layer for both AlexNet and VGG-16 with a novel loss layer matching the target of recovering the high frequencies.

\textbf{Mixed partial derivative loss} Generally, differentiability is an important criteria for studying function space, especially for differential equations, which is the motivation of introducing Sobolev space~\cite{brezis2010functional}.
Roughly speaking, Sobolev space is a vector space equipped with $\mathit{L}_p$ norm \wrt function itself and its weak derivatives.
Motivated by the norms equipped for the Sobolev space, we propose the so-called mixed partial derivative norm for our loss function.
\begin{myrec} 
Let $\alpha = (\alpha_1, \cdots, \alpha_n)$ be the multi-index with $|\alpha| := \alpha_1 + \cdots + \alpha_n$,
and $f$ is $n$-times differentiable function,
then the mixed partial derivative is defined as follows
\begin{equation}
\mathit{D}^{\alpha} f = \frac{\partial^{|\alpha|} f}{\partial x_1^{\alpha_1} \cdots \partial x_n^{\alpha_n}}.
\end{equation}
\end{myrec}
Next we introduce our derivatives norm
\begin{equation}
\label{der}
\|f\|_{k} = \sum_{|\alpha| \leq k} \| \mathit{D}^{\alpha} f \|_2, 
\end{equation}
where 2 indicates the Euclidean norm.
Given that we mainly deal with images and obtain its corresponding derivatives by discrete filters, Eq.~\ref{der} can be converted to the following
\begin{equation}
\label{derd}
\|I\|_{k} = \sum_{|\alpha| \leq k} \| [\mathit{D}]^{\alpha} * I \|_2, 
\end{equation}
where $[\mathit{D}]$ indicates discrete derivative filter and $I$ is the image.
The above formulation is consistent with the preprocessing step by high pass filters (Tab.~\ref{hp}), which are exactly the discrete first- and second derivative operators along $x$ and $y$ direction.
By introducing the mixed partial derivative norm as our loss function, we impose strict constraints on our model so that it decreases the `derivative distances' between denoised output and clean image and keeps the high frequency details. 
In our experiments, we set $k=2$ and ignore the second derivative regarding x and y.  
Since Peak Signal to Noise Ratio (PSNR) is the standard quantitative quality measure of the denoising results, we combine PSNR with the mixed partial derivative norm to obtain the loss function:
\begin{equation}
\label{loss}
\mathit{L}(\mathbf{\Theta}) =  \sum^{|\alpha| \leq 2}_{\alpha \neq (1,1)} -20 \mathbf{log}_{10} \lambda\| [\mathit{D}]^{\alpha} * ( \mathcal{R}(\hat{\mathbf{X}}; \mathbf{\Theta}) +  \hat{\mathbf{X}} - \mathbf{X} )\|_2,
\end{equation}
where $\lambda = \frac{1}{2N}$ and $N$ is the image pixel numbers.

\subsection{Hyperparameters setting}
In order to demonstrate the potential of our model,
we optimize the model hyperparameters by validation. 
To this end, we conduct multiple experiments on the training data provided by ImageNet.
Meanwhile, we randomly select a small subset of validation data from ImageNet.
According to the validation results, we determine important hyperparameters key to the denoising results.

\begin{figure*}[th!]
    \centering    
    \includegraphics[width=0.8\linewidth]{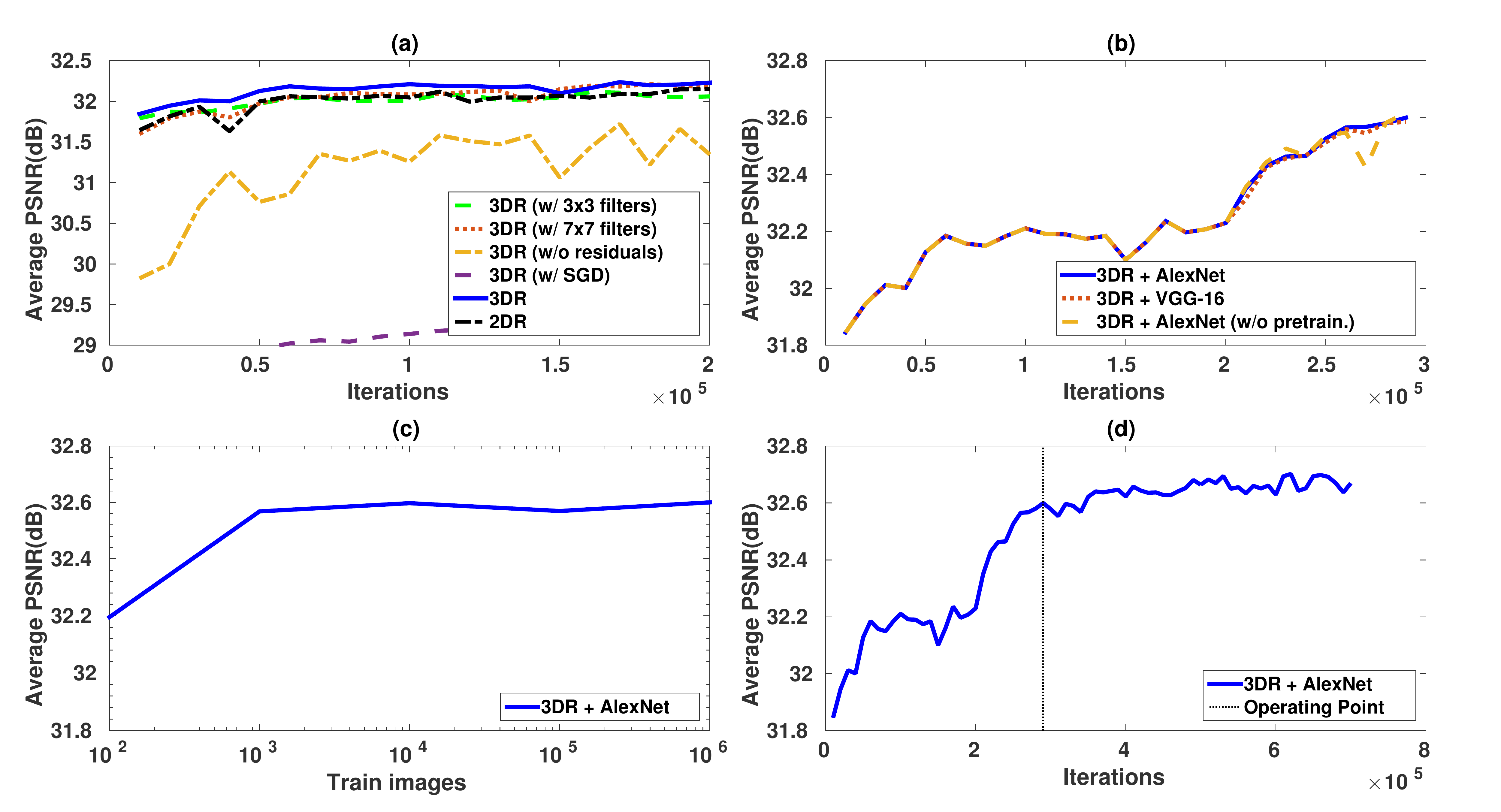}    
    \vspace{-0.2cm}
    \caption{The denoising performances \wrt different hyperparameter configuration.}
    \label{hyper}
    \vspace{-0.3cm}
\end{figure*}

\textbf{Filter size} 
To figure out the spatial size of our 3DR filters we learn 3DR and report the performance with $3\times3$, $5\times5$ and $7\times7$ filter sizes.
As shown in the top left plot of Fig.~\ref{hyper}, our favorite spatial size is $5\times5$.
As shown in Fig.~\ref{hyper} $5\times5$ and $7\times7$ are comparable and both outperform the $3\times3$ configuration. We set our 3DR filters to $5\times5$ size as it is less time to compute than $7\times7$.

\begin{figure*}[th!]
    \centering
    \includegraphics[width=0.7\linewidth]{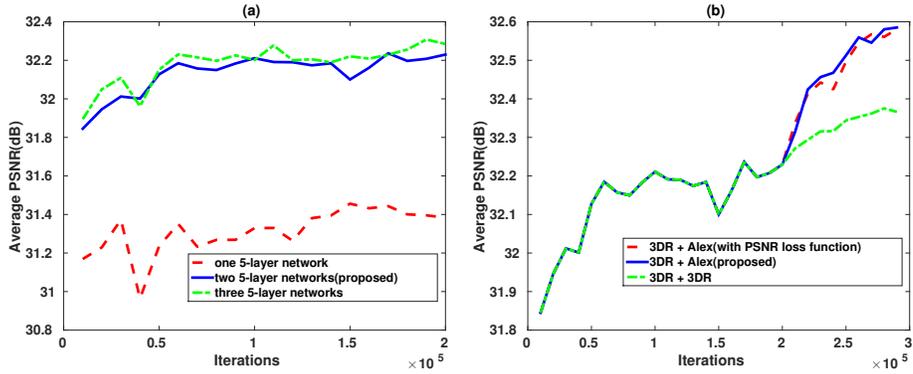}
    \caption{The denoising performances vs. architecture design on the same validation dataset and noise level $\sigma = 15$. (a) shows how the number of redundant 5-layer network architecture design affects average PSNR, we compare the cases of one 5-layer network, the combination of two and three 5-layer network; (b) shows how loss function and depth of 3DR affects average PSNR, we compare the case of using PSNR and mixed partial derivative loss, and cascading 3DR twice. }
    \label{fig:fig1}
    \vspace{-0.3cm}
\end{figure*}

\textbf{Residual vs. Image}
Residual learning is a useful trick to overcome the accuracy saturation problem,
demonstrating its strength in various tasks including image super-resolution~\cite{kim2015accurate}.
Our experiments on denoising task substantiate the effectiveness of residual learning,
Fig.~\ref{hyper} presents an obvious PSNR gap between working on residuals and when using the image itself.

\textbf{AlexNet and VGG-16}
We verify the denoising performance obtained by AlexNet and VGG-16.
During the training of second stage, we simply follow their original parameter setup and freeze the 3RD weights. 
We also compare AlexNet with finetuning of the pretrained ImageNet model (for classification) and AlexNet with training from scratch.
Experimental results in Fig.~\ref{hyper}(b) show that both of them achieve competitive results and boost the results with a large margin over the first stage. Using the pretrained weights with finetuning seem to lead to a more stable training.
Therefore, by default we apply the finetuning strategy for both AlexNet and VGG-16.

\textbf{Training}
For training we use Adam~\cite{kingma2014adam} instead of the commonly used stochastic gradient descent (SGD). With Adam we achieve a rapid loss decrease in the first several hundreds iterations of the first 3DR stage and further accuracy improvement during the second stage. With SGD we achieve significantly poorer results than with Adam.
By cross validation we determine the learning rate to be $0.005, 0.001$ for the first and second stage, and momentum to be $0.9$ throughout the stage. 
The minibatch size is set to 10 and the number of iterations are set to be $200000, 90000$ accordingly for the first and second stage. 
100 training image brings the performance to 32.23dB (see Fig.~\ref{hyper}(c), slightly better than the first stage result and indicates a clear overfitting and saturation in performance. Increasing the train pool of images to 1000 brings a consistent $\sim0.4dB$ improvement close to the maximum performance achieved when using 1 million train images.
We conclude that for denoising it is necessary to use large sets for training, in the range of thousands. Though the improvement margin is relative small between using 1000 and 1000000 train images, it is no harm to train on million of images for increased robustness.
While for all the experimental results we kept the models trained with 290000 iterations (due to lack of time and computational resources) we point out that the performance of our models is still improving with the extra number of iterations ($\sim0.1dB$ in Fig.~\ref{hyper}(d) for 500000 iterations over the 290000 iterations operating point).

\begin{figure*}[th!]
\centering
\resizebox{1.0\linewidth}{!}
{
\begin{tabular}{ccccc}
\huge input &\huge TRND &\huge CBM3D &\huge \textbf{Ours} &\huge ground truth\\
\includegraphics[width=0.4\linewidth]{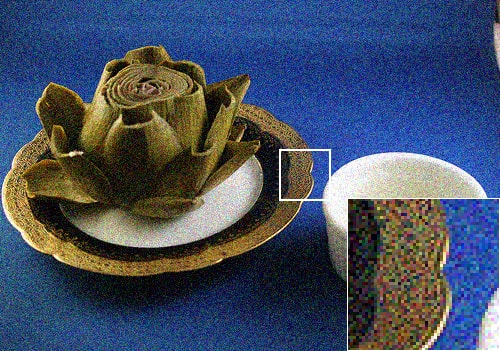}&
\includegraphics[width=0.4\linewidth]{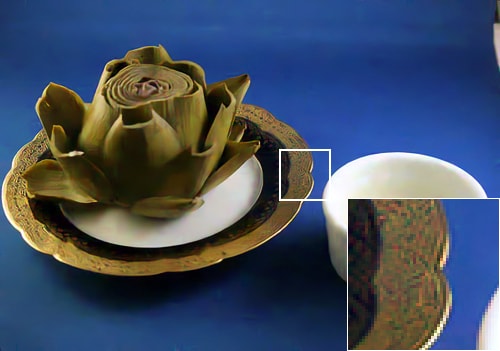}&
\includegraphics[width=0.4\linewidth]{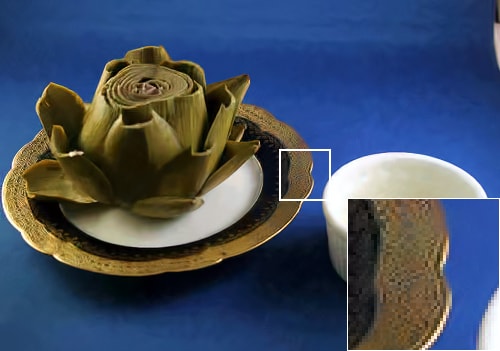}&
\includegraphics[width=0.4\linewidth]{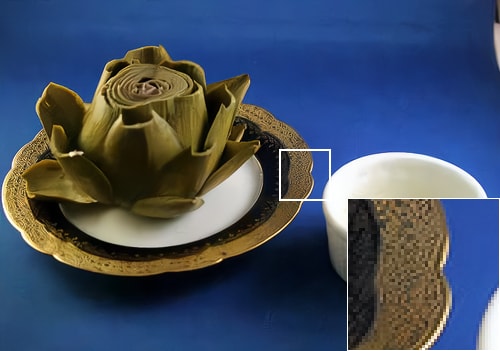}&
\includegraphics[width=0.4\linewidth]{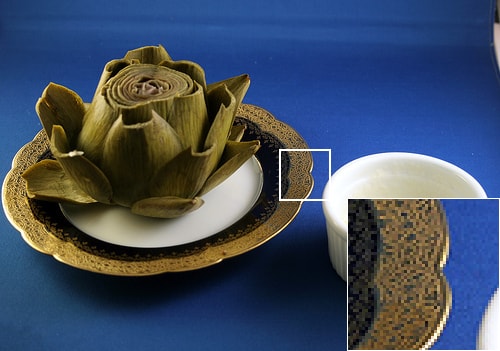}\\

\includegraphics[width=0.4\linewidth]{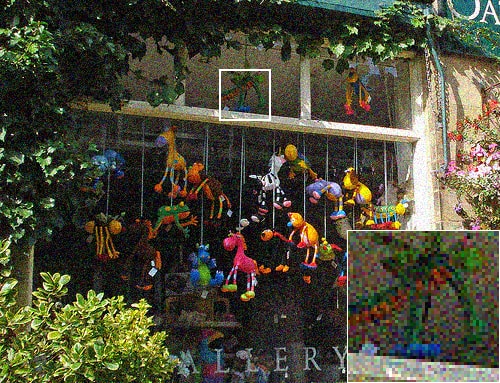}&
\includegraphics[width=0.4\linewidth]{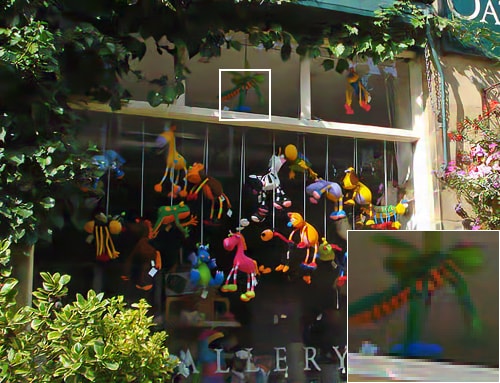}&
\includegraphics[width=0.4\linewidth]{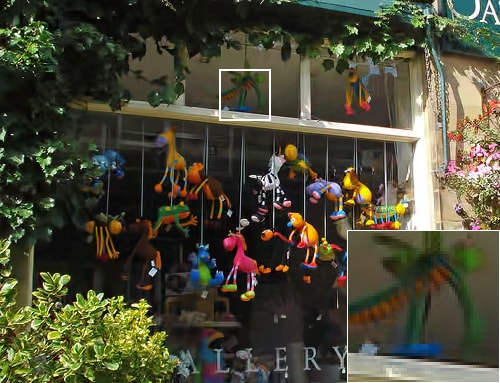}&
\includegraphics[width=0.4\linewidth]{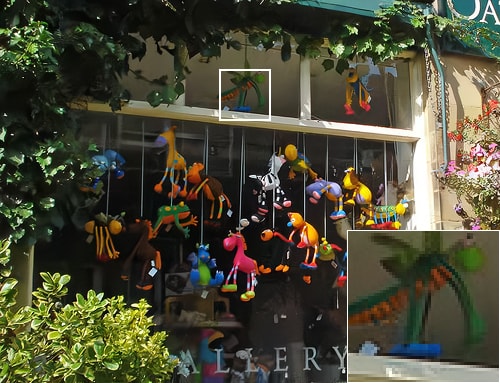}&
\includegraphics[width=0.4\linewidth]{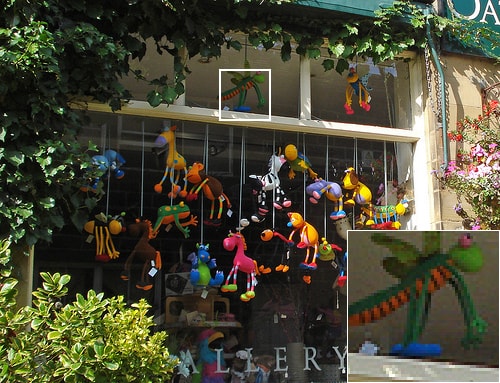}\\
\end{tabular}
}       
\caption{Visual assessment of non-blind denoising results for $\sigma=25$ on ImageNet validation. Best zoom on screen.}
\label{fig:n25}
\vspace{-0.2cm}
\end{figure*}
\begin{figure*}[th!]
\centering
\setlength{\tabcolsep}{1pt}
\resizebox{\linewidth}{!}
{
\begin{tabular}{cccccc}
\huge input &\huge  TRND & CBM3D &\huge  HOSVD+Wiener &\huge  \textbf{Ours} & \huge ground truth\\

\includegraphics[width=0.2\linewidth]{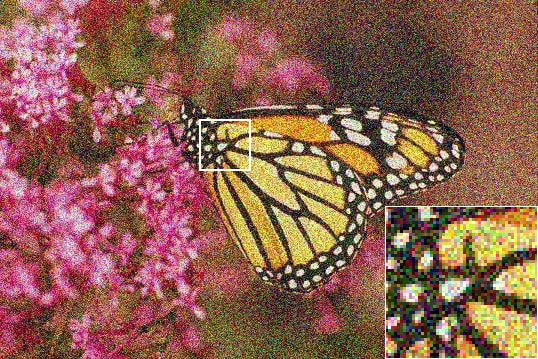}&
\includegraphics[width=0.2\linewidth]{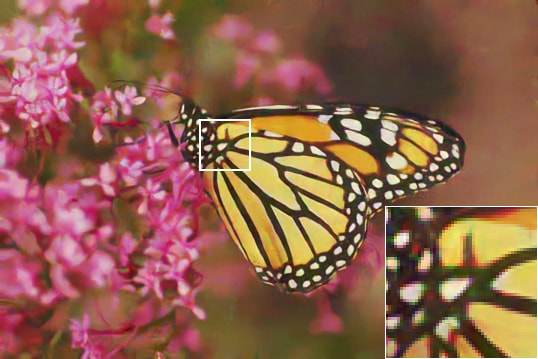}&
\includegraphics[width=0.2\linewidth]{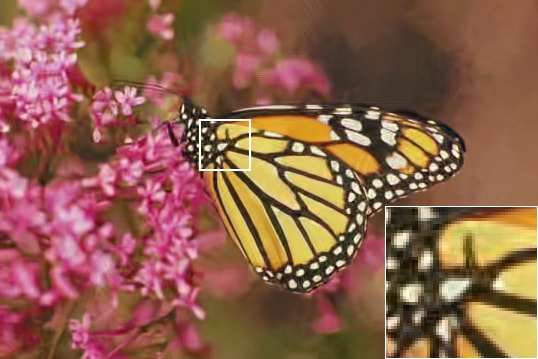}&
\includegraphics[width=0.2\linewidth]{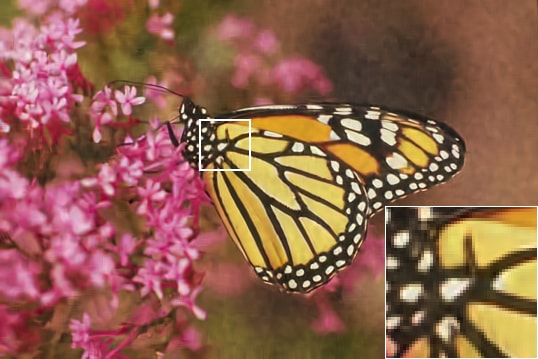}&
\includegraphics[width=0.2\linewidth]{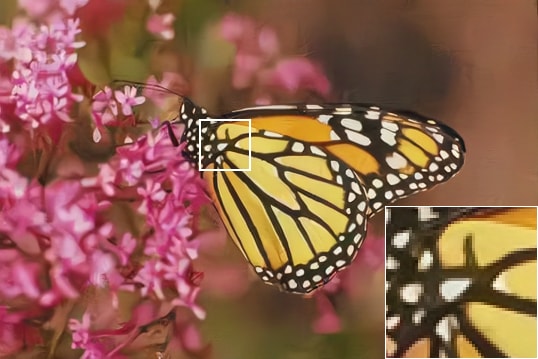}&
\includegraphics[width=0.2\linewidth]{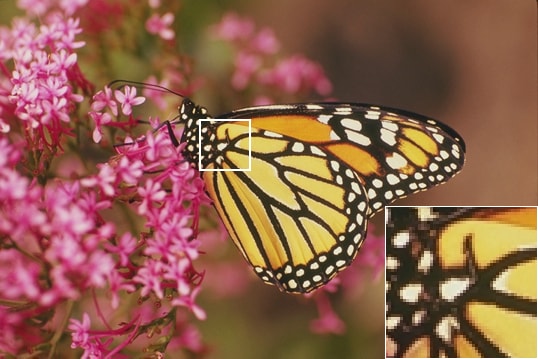}\\
\includegraphics[width=0.2\linewidth]{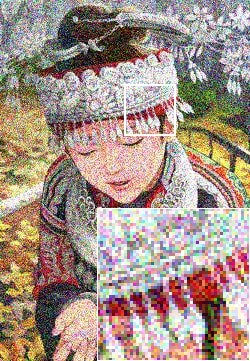}&
\includegraphics[width=0.2\linewidth]{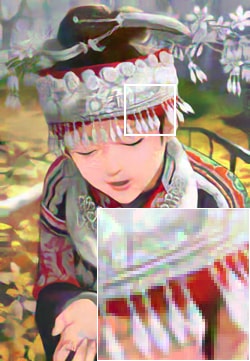}&
\includegraphics[width=0.2\linewidth]{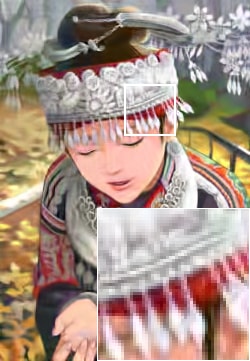}&
\includegraphics[width=0.2\linewidth]{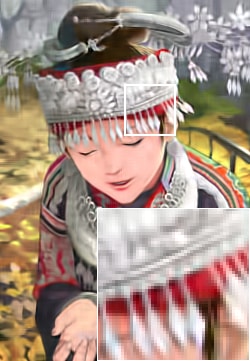}&
\includegraphics[width=0.2\linewidth]{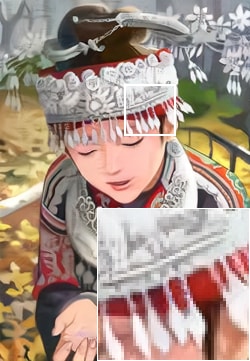}&
\includegraphics[width=0.2\linewidth]{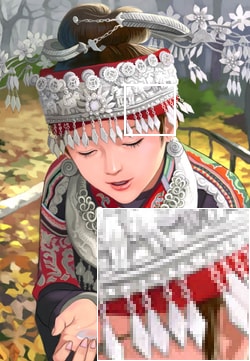}\\
\end{tabular}
}       
    \caption{Visual assessment of non-blind denoising results for $\sigma=50$ on Set 14. Best zoom on screen.}
    \label{fig:n50}
\vspace{-0.2cm}    
\end{figure*}

\begin{figure*}[th!]
\centering
\setlength{\tabcolsep}{1pt}
\resizebox{\linewidth}{!}
{
\begin{tabular}{ccccc}
input & TRND & CBM3D & \textbf{Ours} & ground truth\\
\includegraphics[width=0.4\linewidth]{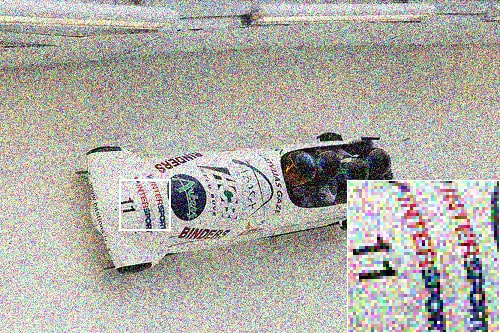}&
\includegraphics[width=0.4\linewidth]{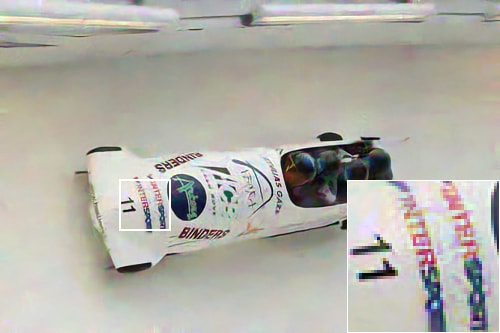}&
\includegraphics[width=0.4\linewidth]{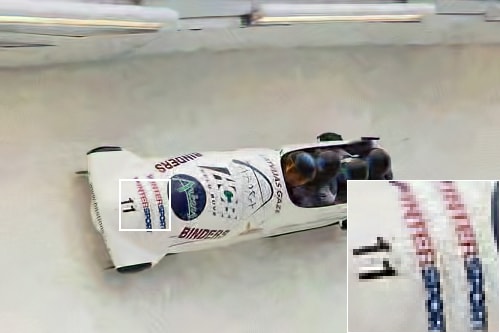}&
\includegraphics[width=0.4\linewidth]{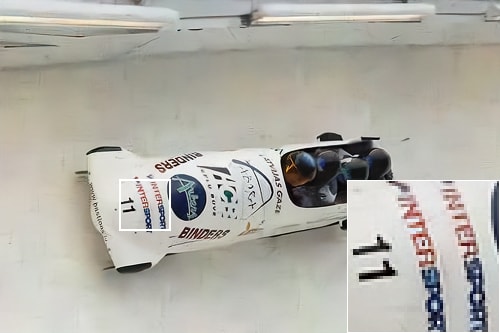}&
\includegraphics[width=0.4\linewidth]{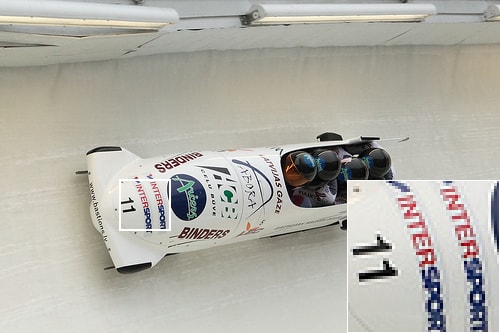}\\
\includegraphics[width=0.4\linewidth]{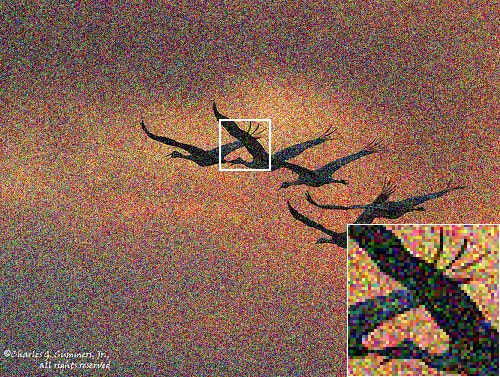}&
\includegraphics[width=0.4\linewidth]{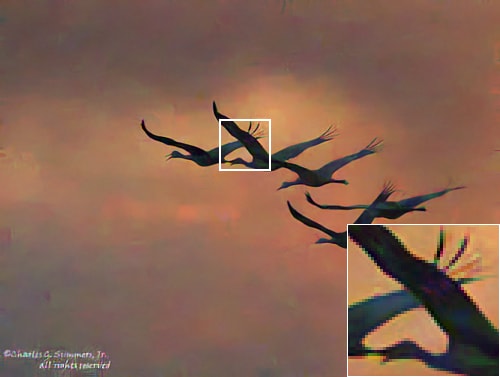}&
\includegraphics[width=0.4\linewidth]{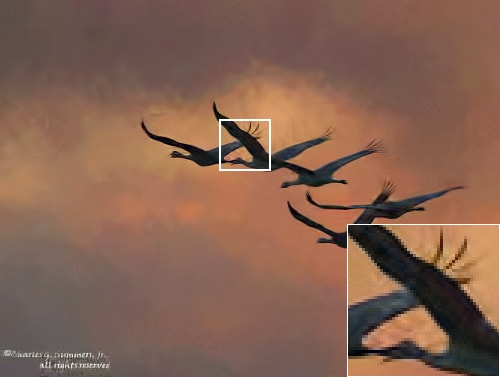}&
\includegraphics[width=0.4\linewidth]{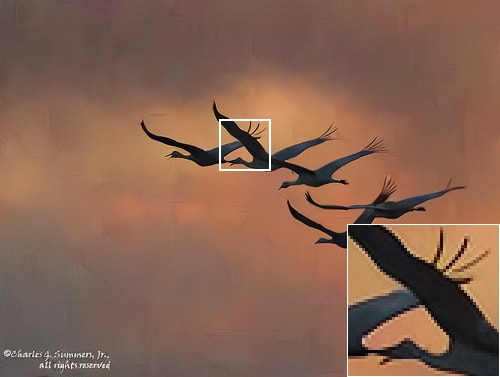}&
\includegraphics[width=0.4\linewidth]{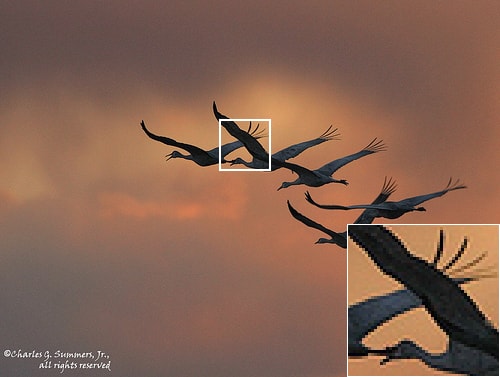}\\
\includegraphics[width=0.4\linewidth]{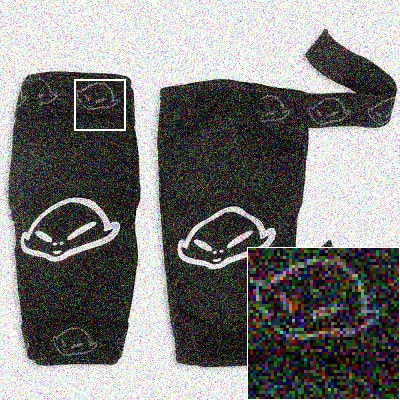}&
\includegraphics[width=0.4\linewidth]{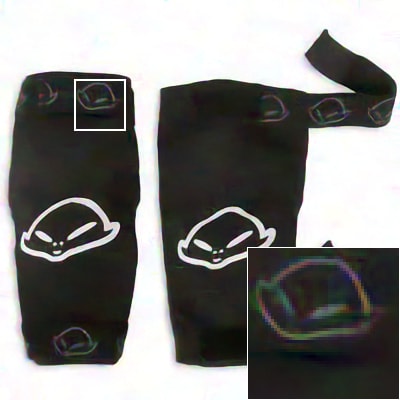}&
\includegraphics[width=0.4\linewidth]{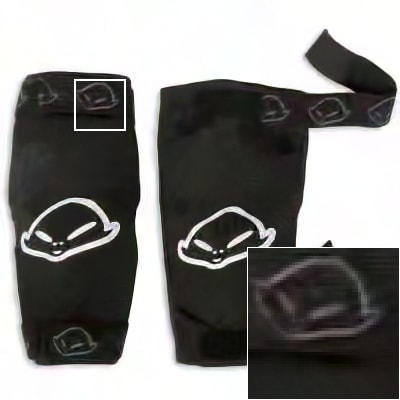}&
\includegraphics[width=0.4\linewidth]{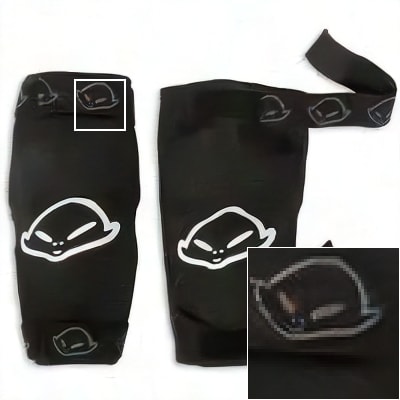}&
\includegraphics[width=0.4\linewidth]{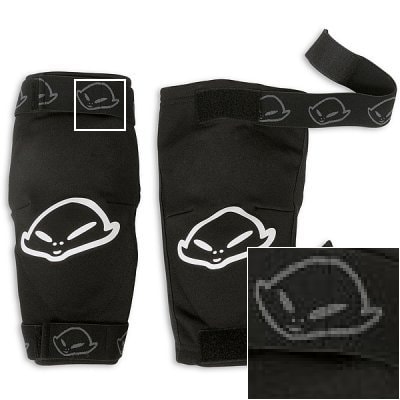}\\

\end{tabular}
}       
    \caption{Visual assessment of non-blind denoising results for $\sigma=50$ on ImageNet. Best zoom on screen.}
    \label{fig:n50I}
\vspace{-0.2cm}    
\end{figure*}
\subsection{Architecture Design Choices}
In this section, 
we report a couple of experimental results which support the architecture decisions we made.

\textbf{Architecture Redundancy}
Firstly, for 3DR we picked up a design redundancy of two 5-layer networks for which we then average their residuals.
In Fig.~\ref{fig:fig1}(a) we report the effect on the PSNR performance of 3DR with 1, 2, and 3 such 5-layer networks on the same validation data(\ie small subset of validation images in ImageNet) and noise level(\ie $\sigma=15$) mentioned in our paper. The three 5-layer networks architecture of 3DR leads to the best performance for most cases of training iterations. The one 5-layer architecture is much worse than the two 5-layer.

If we apply three 5-layer, we can still slightly improve the PSNR over the two 5-layer design, however the performance gain begins to saturate at the cost of extra computational time. Therefore the two 5-layer networks are the default setting for our 3DR model and if sufficient time and memory budget is available then by increasing the number of 5-layer networks in the redundant design of 3DR some improvements are clearly possible.

\textbf{Architecture Components}
As analyzed above we propose a 3DR component which can be trained separately and embeds the 3D filtering and joint processing of the color image channels.
On top of this standalone component we can cascade either other 3DR components or components derived from published architectures such as AlexNet and VGG-16 proved successful for classification tasks.
In Fig.~\ref{fig:fig1}(b) we report the results on the same dataset and noise level as Fig.~\ref{fig:fig1}(a) when we stack on top of our first 3DR component either: i) another 3DR component or ii) the adapted AlexNet architecture as described in the paper.
In both cases we benefit from deepening our network design. However, 
AlexNet is deeper than 3DR and with fewer training iterations leads to significantly better PSNR performance. 
Thus, it is helpful to introduce a deep architecture (from classification) as our second stage.
Of course, with enough computation resources,
we can still gain PSNR improvement by cascading 3DR multiple times or stacking more layers in 3DR.

\textbf{Loss Function}
When it comes to the loss function, we prefer our mixed partial derivative (Eq.~\ref{loss}) loss to a PSNR loss.
In Fig.~\ref{fig:fig1}(b) we compare the PSNR performance of our proposed architecture (3DR+AlexNet) when using the mixed partial derivative loss and when using the PSNR on the second stage/component -- AlexNet. For the first 3DR stage we always use PSNR loss.
We consider the two stages of our model to be coarse to fine,
thus during the coarse stage we use the normal PSNR loss.
A mild improvement is achieved when using the mixed partial derivative loss over PSNR loss shown in Fig.~\ref{fig:fig1}(b).


\section{Experiments}
\label{sec:experiments}

\subsection{Image denoising}
Our experiments are mainly conducted on the large ImageNet dataset. 
We train our models with the complete ImageNet training data containing more than 1 million images.
Due to the GPU memory limitation and for training efficiency, we randomly crop the training image with size $180\times180$ and set minibatch to be $10$. 
For testing we have 1000 images with original size, collected from the validation images of ImageNet, which have not been used for the hyperparameter validation.
Complementary to the ImageNet test images we also test our methods on traditional denoising and reconstruction benchmarks: Kodak and Set14~\cite{zeyde2010single,timofte2013anchored}, which contains 24 and 14 images, resp. Set14 contains classical images such as Lena, Barbara, Butterfly.

\begin{table}[th!]
\centering
\caption{The average PSNR and SSIM performances for Kodak dataset. The best results are in bold.}
\label{kodak}
\resizebox{\columnwidth}{!}
{
\begin{tabular}{ccccccccccc}
\hline
\multirow{3}{*}{Methods} & \multicolumn{10}{c}{$\sigma$}                                                                                                            \\ \cline{2-11} 
                         & \multicolumn{2}{c}{15}                      & \multicolumn{2}{c}{25}                      & \multicolumn{2}{c}{50}  
                         & \multicolumn{2}{c}{90}                      & \multicolumn{2}{c}{130}\\ \cline{2-11} 
                         & PSNR                 & SSIM                 & PSNR                 & SSIM                 & PSNR                 & SSIM  
                         & PSNR                 & SSIM                 & PSNR                 & SSIM \\ \hline
TRND                     & 32.48                & 0.9420               & 30.10                & 0.9085               & 27.23                & 0.8464 
                         &- &- &- &-\\
CBM3D                    & 34.43                & 0.9621               & 31.83                & 0.9377               & 28.63                & 0.8865 
                         & 26.27&0.8288 &23.48 &0.6929\\
HOSVD                    & 33.66 & 0.9540 & 31.12 & 0.9248 & 27.97 & 0.8674 &- &- &- &- \\
HOSVD+wiener             & 34.18 & 0.9594 & 31.64 & 0.9332 & 28.38 & 0.8781 &- &- &- &- \\ \hline
3DR+VGG-16(b)                 & 33.99                & 0.9587               & 31.84                & 0.9373               & 28.64                & 0.8824               
                         &19.87 &0.4724 &- &-\\
3DR+AlexNet(b)                  & 33.98       & 0.9587      & 31.87       & 0.9380      & 28.72       & 0.8856     
                         &19.40 &0.4975 &- &-\\ 
3DR+VGG-16                 & 34.66                & 0.9635               & 32.11                & 0.9403               & 28.88                & 0.8891               
                         &26.37 &0.8243 &24.85 &0.7723\\
3DR+AlexNet                  & 34.69       & 0.9636      & 32.16       & 0.9409      & 28.98       & 0.8906     
                         &26.46 &0.8282 &25.00 &0.7789\\ 
3DR+VGG-16(r)                 & 34.75                & 0.9641               & 32.20                & 0.9413               & 28.98                & 0.8912               
                         &26.48 &0.8282 &24.97 &0.7772\\
3DR+AlexNet(r)                  & \textbf{34.79}       & \textbf{0.9643}      & \textbf{32.27}       & \textbf{0.9423}      & \textbf{29.06}       & \textbf{0.8932}      
                         &\textbf{26.61} &\textbf{0.8335} &\textbf{25.15} &\textbf{0.7860}\\ 
\hline
\end{tabular}
}
\vspace{-0.2cm}
\end{table}

\begin{table}[th!]
\centering
\caption{The average PSNR and SSIM performances for Set14 dataset. The best results are in bold.}
\label{lena}
\resizebox{\columnwidth}{!}
{
\begin{tabular}{ccccccccccc}
\hline
\multirow{3}{*}{Methods} & \multicolumn{10}{c}{$\sigma$}                                                                                                            \\ \cline{2-11} 
                         & \multicolumn{2}{c}{15}                      & \multicolumn{2}{c}{25}                      & \multicolumn{2}{c}{50}  
                         & \multicolumn{2}{c}{90}                      & \multicolumn{2}{c}{130} \\ \cline{2-11} 
                         & PSNR                 & SSIM                 & PSNR                 & SSIM                 & PSNR                 & SSIM 
                         & PSNR                 & SSIM                 & PSNR                 & SSIM\\ \hline
TRND                     & 31.88                & 0.9450               & 29.54                & 0.9125               & 26.49                & 0.8472            
                         &- &- &- &-\\
CBM3D                    & 33.18                & 0.9610               & 30.84                & 0.9375               & 27.83                & 0.8875               
                         &25.31 &0.8267  &22.67 &0.7211\\
HOSVD                    & 32.77 & 0.9553 & 30.51 & 0.9285 & 27.34 & 0.8706 &- &- &- &-\\
HOSVD+wiener             & 33.15 & 0.9596 & 30.91 & 0.9360 & 27.77 & 0.8834 &- &- &- &-\\ \hline
3DR+VGG-16(b)                 & 32.34                & 0.9539               & 30.69                & 0.9352               & 27.76                & 0.8864               
                         &19.67 &0.6079 &- &-\\
3DR+AlexNet(b)                  & 32.33       & 0.9538      & 30.71       & 0.9356      & 27.85       & 0.8887     
                         &19.32 &0.5929 &- &-\\
3DR+VGG-16                  & 33.38                & 0.9624               & 31.05                & 0.9394               & 27.97                & 0.8901               
                         &25.29 &0.8248  &23.68  &0.7739 \\
3DR+AlexNet                 & 33.41       & 0.9625      & 31.11       & 0.9400      & 28.01       & 0.8904 
                         &25.38 &0.8271 &23.81 &0.7776\\
3DR+VGG-16(r)                  & 33.48                & 0.9631               & 31.18                & 0.9407               & 28.10                & 0.8923               
                         &25.46 &0.8292 &23.86 &0.7798\\
3DR+AlexNet(r)                 & \textbf{33.53}       & \textbf{0.9634}      & \textbf{31.26}       & \textbf{0.9416}      & \textbf{28.17}       & \textbf{0.8932}  
                         &\textbf{25.58} &\textbf{0.8330} &\textbf{24.01} &\textbf{0.7852}\\ \hline
\end{tabular}
}
\vspace{-0.2cm}
\end{table}

\begin{table}[th!]
\centering
\caption{The average PSNR and SSIM performances for 1000 images from validation data of ImageNet. The best results are in bold.}
\label{imagenet}
\resizebox{\columnwidth}{!}
{
\begin{tabular}{ccccccccccc}
\hline
\multirow{3}{*}{Methods} & \multicolumn{10}{c}{$\sigma$}                                                                           \\ \cline{2-11} 
                         & \multicolumn{2}{c}{15}           & \multicolumn{2}{c}{25}           & \multicolumn{2}{c}{50}           
                         & \multicolumn{2}{c}{90}           & \multicolumn{2}{c}{130}\\ \cline{2-11} 
                         & PSNR           & SSIM            & PSNR           & SSIM            & PSNR           & SSIM 
                         & PSNR           & SSIM            & PSNR           & SSIM\\ \hline
TRND                     & 31.54          & 0.9328          & 29.11          & 0.8951          & 26.12          & 0.8231
                         &-&-&-&-\\
CBM3D                    & 32.98          & 0.9504          & 30.47          & 0.9215          & 27.30          & 0.8604          
                         &24.94  &0.7930  &22.54  &0.6672\\
3DR+VGG-16                 & 33.29          & 0.9528          & 30.86          & 0.9262          & 27.76          & 0.8710          
                         &25.27  &0.8017  &23.82  &0.7499\\
3DR+AlexNet                  & \textbf{33.32} & \textbf{0.9530} & \textbf{30.92} & \textbf{0.9269} & \textbf{27.81} & \textbf{0.8713} 
                         &\textbf{25.35}  &\textbf{0.8041}  &\textbf{23.94}  &\textbf{0.7539}\\ \hline
\end{tabular}
}
\vspace{-0.2cm}
\end{table}
\textbf{Non-blind denoising}
Most typical setup is when the noise is Gaussian with known standard deviation $\sigma$. 
Given a wide range of $\sigma = 15, 25, 50, 90, 130$ we compare our methods with CBM3D, recently proposed HOSVD~\cite{rajwade2013image}~\footnote{Unfortunately, due to high time complexity of HOSVD we did not conduct its experiments on ImageNet.} and its variant HOSVD with wiener filter (HOSVD+Wiener), as well as one state-of-the-art grayscale image denoising method TRND~\cite{chen2015learning} applied independently on each image channel.
The performance on Set14, Kodak and ImageNet datasets is reported in Tab.~\ref{lena},~\ref{kodak}, and ~\ref{imagenet}.
Our proposed methods reach the best performances for all testing datasets.
Fig.~\ref{hyper}(a) reports 3DR performance and Fig.~\ref{hyper}(b) shows further gains with
AlexNet/VGG-16. 
Generally the full 3DR+AlexNet/VGG-16 method are
0.2-0.4dB higher than 3DR.
On ImageNet our 3DR+AlexNet goes from an improvement of 0.34dB PSNR over CM3D for small levels of noise ($\sigma=15$) to 0.51dB for medium levels of noise ($\sigma=50$) and to 1.5dB for high noise ($\sigma=130$). 
The larger the noise ($\sigma$) is, the larger the performance gap gets. Thus, our methods cope well with high levels of noise while for CBM3D it is getting more difficult to correctly group similar patches. 3DR+AlexNet performs sligtly better than 3DR+VGG-16.
Our models perform the best also when using the SSIM measure, which in our experiments correlates well with PSNR.
On Kodak and Set14 the denoising results and the relative performance are consistent with the ones on ImageNet. Considering also that our methods were trained on ImageNet train dataset which has a different distribution than Kodak or Set14 we can conclude that our methods, 3DR+AlexNet/VGG-16, do not overfit ImageNet images and generalizes well for color images outside this large dataset.
Tab.~\ref{kodak} and~\ref{lena} also shows that our methods are better than the recent HOSVD methods, which cost hours to process a single image, and that using the inter-channel correlations for color image denoising is a must, otherwise top state-of-the-art single channel denoising methods such as TRND provide poor reference performance.
In addition, we use enhanced prediction trick~\cite{Timofte-CVPR-2016} and rotate the noisy images by 0, 90, 180, 270 degrees, process them and then rotate back and average the denoised outputs. In this way we achieve significant performance gains for both our methods 3DR+AlexNet(e) and 3DR+VGG-16(e) at the cost of increased running time. For instance, on Set14, 3DR+AlexNet(e) gains 0.16dB over 3DR+AlexNet at $\sigma=50$ and 0.2dB at $\sigma=130$. 
Visual results also confirm the quantitative improvements achieved by our methods (See Fig.~\ref{fig:n25},~\ref{fig:n50},~\ref{fig:n50I})

\begin{figure}[ht!]
    \centering
    \includegraphics[width=\columnwidth]{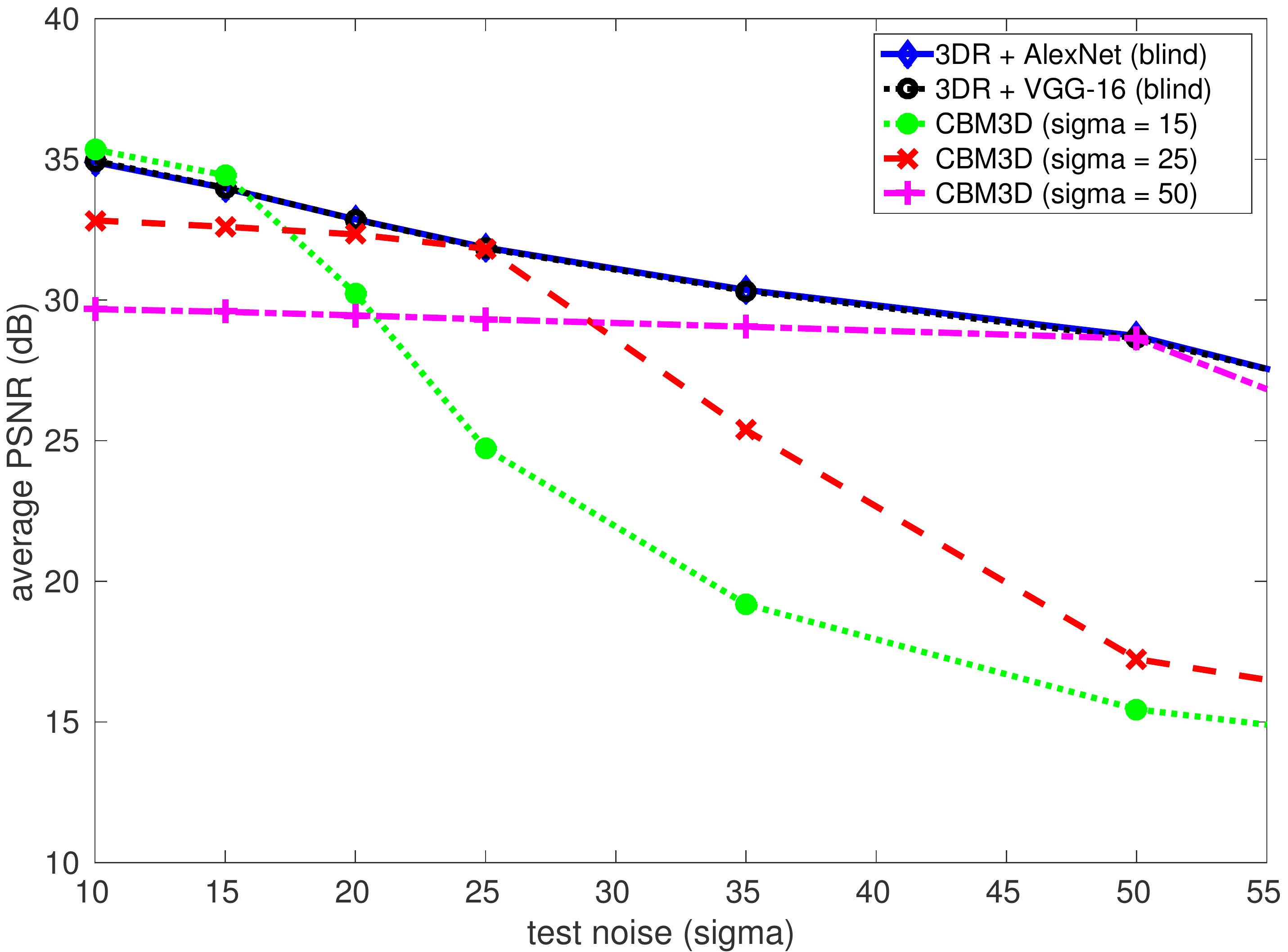}
    \caption{Blind denoising comparison of our 3DR+AlexNet and 3DR+VGG-16 blindly trained methods and CBM3D with fixed noise levels $\sigma\in\{15,25,50\}$}
    \label{fig:blind_denoising}
\end{figure}

\textbf{Blind denoising}
In practice, the noise level is seldom known and can vary with the source of noise and time. Therefore, the methods are able to cope `blindly' with unknown levels of noise are very important for real applications.
As shown in Fig.~\ref{fig:blind_denoising} for the reference CBM3D method whenever there is a mismatch between the level of noise in the test image and the one the method is set for the performance significantly degrades in comparison with the same method with `knowledge' of the noise level at test.

In order to test the robustness of our proposed denoising methods (3DR+AlexNet and 3DR+VGG-16) to blind denoising conditions we train using the same settings as before, for the non-blind denoising case. The only difference is that the noise level randomly changes from one train image sample to another during the training. In this way the deep model learns to denoise Gaussian noise `blindly'. In Fig.~\ref{fig:blind_denoising} we compare the performance of our blind methods with that of the CBM3D method with knowledge of 3 levels of noise, $\sigma\in\{15,25,50\}$. The test images are the 24 Kodak images. Our blind methods perform similarly regardless the level of noise and are effective envelopes for the performance achieved using CBM3D with various noise settings. Note that only for the low levels of noise ($\sigma<15$) our blind methods perform poorer than the non-blind CBM3D. At the same time, as expected, our blind models (3DR+VGG-16(b), 3DR+AlexNet(b)) achieve a denoising performance below the one of our non-blind models as numerically shown in Tab.~\ref{lena} and~\ref{kodak}.

\begin{figure}
\centering
\setlength{\tabcolsep}{1pt}
\begin{tabular}{ccc}
Real Noise & 3DR+AlexNet(b) & CBM3D \\
\includegraphics[width=0.32\linewidth]{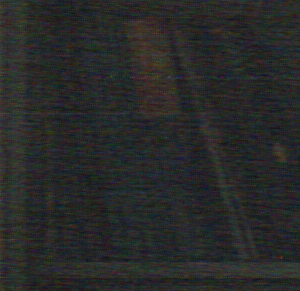}&
\includegraphics[width=0.32\linewidth]{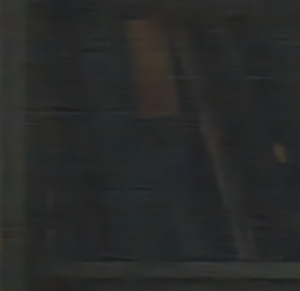}&
\includegraphics[width=0.32\linewidth]{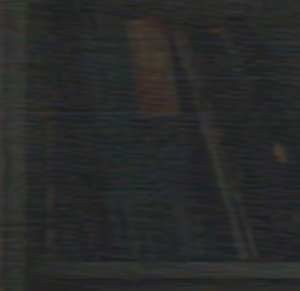}\\
\end{tabular} 
\caption{Real image denoising, the best CBM3D result is shown. Best zoom on screen.}
\label{fig:real}
\vspace{-0.3cm}
\end{figure}

We also present here a visual result for real image denoising Fig.~\ref{fig:real},
which means we have no clue about the noise pattern. 
The image was taken under dim light conditions with truly poor image quality. 
Visual result shows that our method still provides better image quality compared to the best CBM3D result.

\textbf{Running time}
We use the GeForce GTX TITAN X GPU card to train and test our models.
It takes about one day for the 200,000 iterations of the 3DR training phase.
Another 1.5 days are necessary for the following 90,000 iterations to train a whole model (3DR+AlexNet or 3DR+VGG-16). At test, an image with size $500\times 375$ is processed in about 60 seconds, including CPU and GPU memory transfer time. In comparison, HOSVD costs about 4 hours to denoise the same image on CPU.

\textbf{Visualization}
We visualize the first layer filters of 3DR and finetuned AlexNet (See Fig.~\ref{fig:3dr} and~\ref{fig:alex}) to gain some insights of proposed model. 
Filters on noisy image itself (middle plot of Fig.~\ref{fig:3dr}) shows that color is useful for denoising, 
there exist green and pink filter among other edge filters.
Moreover, the filters obtained by finetuning ImageNet (Fig.~\ref{fig:alex}) seems quite interesting. 
We can easily notice that most of the finetuned filters (2-4 plot of Fig.~\ref{fig:alex})share lots in common with the original classification filters of AlexNet. 
It suggests that filters trained for high level task can also help the low level task such as denoising.
There are only few filters with white background which are completely different from the original ones, 
we believe those filters are particularly adapted to detect Gaussian noise.

\begin{figure}[bt!]
\begin{tabular*}{\columnwidth}{@{}c|cccc@{}}
Example & Category & general training & Category training \\
\includegraphics[width=0.7cm, height=0.7cm]{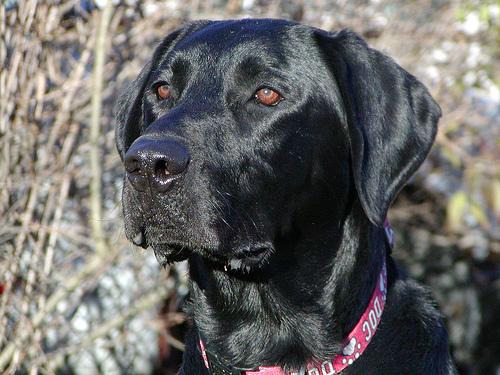}
& Labrador & 33.38 & \textbf{33.44}\\

\includegraphics[width=0.7cm, height=0.7cm]{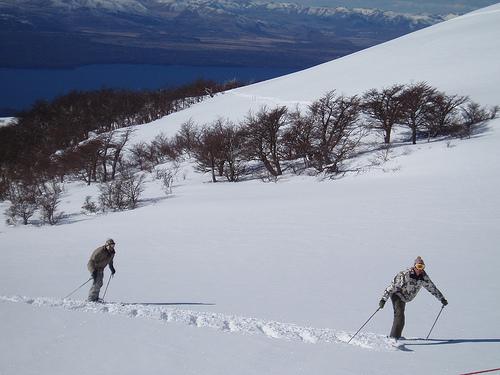}
& Ski & 33.41 & \textbf{33.46}\\

\includegraphics[width=0.7cm, height=0.7cm]{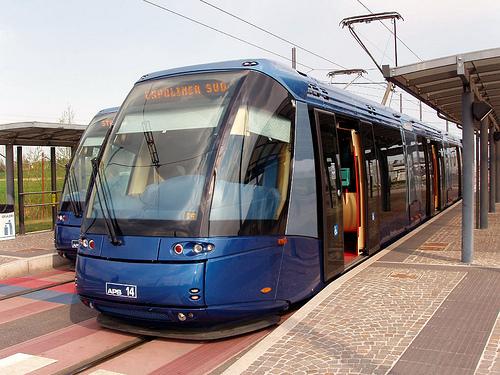}
& Tram & 32.13 & \textbf{32.30}\\

\includegraphics[width=0.7cm, height=0.7cm]{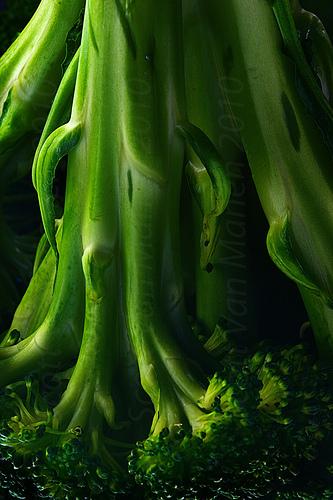}
& Broccoli & 31.62 & \textbf{31.67}\\

\includegraphics[width=0.7cm, height=0.7cm]{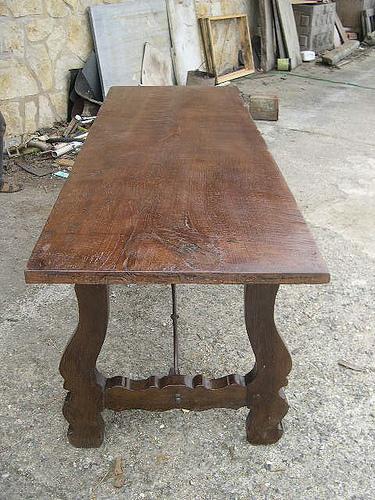}
& Dining table & 33.45 & \textbf{33.47}\\

\includegraphics[width=0.7cm, height=0.7cm]{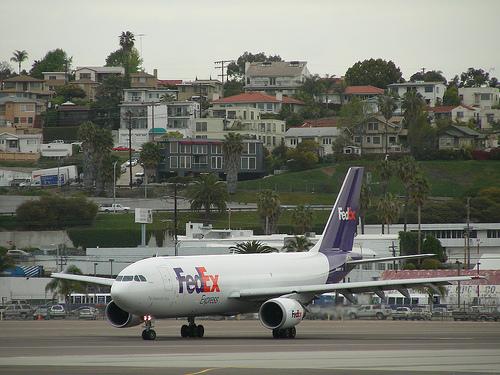}
& Airliner & 34.49 & \textbf{34.62}\\

\end{tabular*}
\caption{PSNR performances achieved by generic - and semantic label finetuning.}
\label{Fig:erp}
\vspace{-0.25cm}
\end{figure}

\begin{figure}
\setlength{\tabcolsep}{1pt}
\begin{tabular}{c|ccc|ccc}
class & \multicolumn{3}{c|}{Airliner} & \multicolumn{3}{c}{Dining table} \\
image & \multicolumn{3}{c|}{\includegraphics[width=0.37\columnwidth]{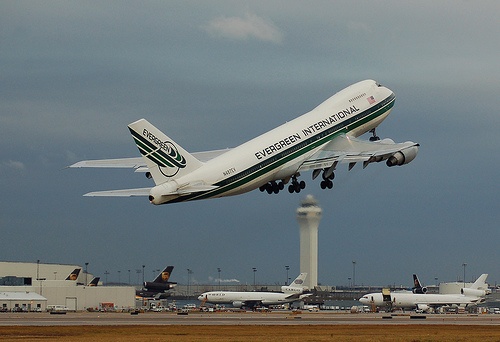}}&\multicolumn{3}{c}{\includegraphics[width=0.37\columnwidth]{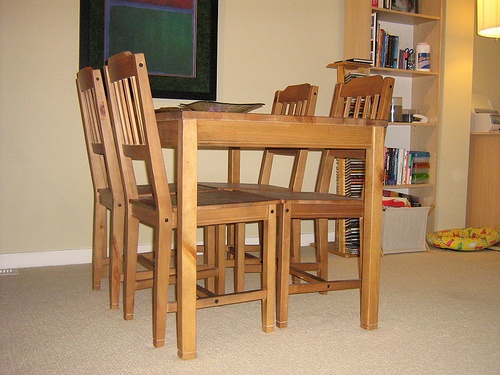}}\\
mask  & \includegraphics[width=0.11\columnwidth]{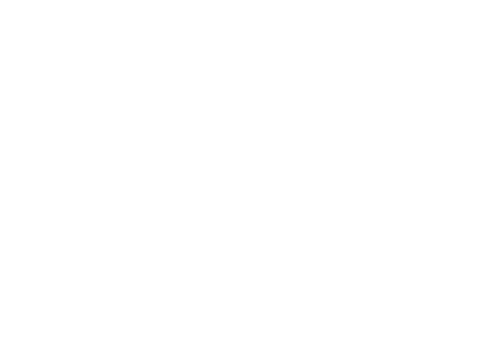}&\includegraphics[width=0.11\columnwidth]{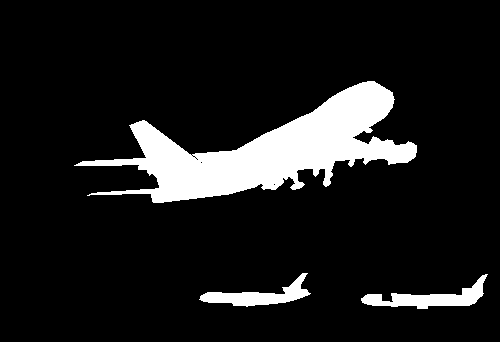}&\includegraphics[width=0.11\columnwidth]{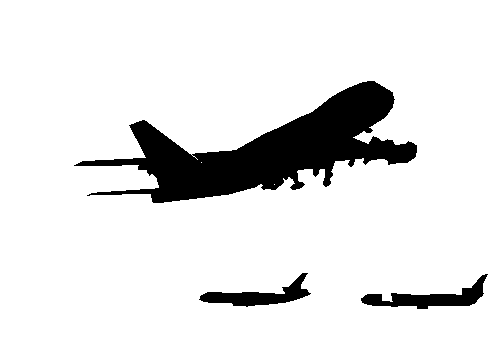} & \includegraphics[width=0.11\columnwidth]{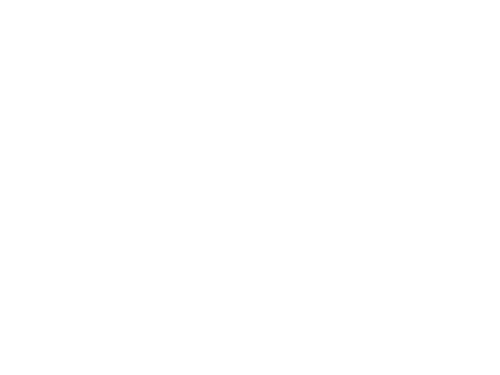}&\includegraphics[width=0.11\columnwidth]{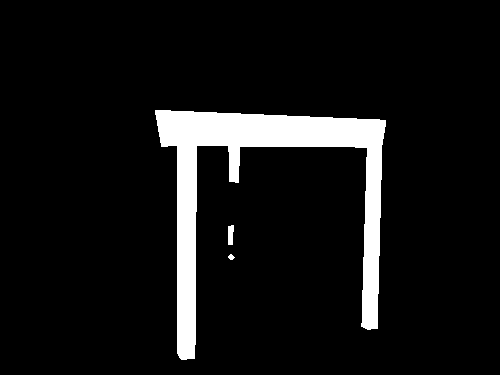}&\includegraphics[width=0.11\columnwidth]{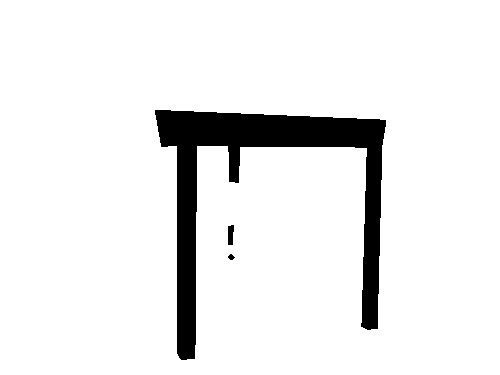}\\
\hline
generic & 36.60& 32.53 & 38.36 & 34.21 & 33.62 & 34.77\\
semantic& 36.66& 32.62 & 38.37 & 34.24 & 33.67 & 34.79\\
\hline
improvement&+0.06&+0.09&+0.01&+0.03&+0.07&+0.02\\
\end{tabular}
\caption{Denoising PSNR(dB) results for two examples when 3DR + AlexNet is trained using generic settings (no semantic knowledge) and when is trained for each semantic class individually. $\sigma=15$}
\label{fig:semantic_examples}
\end{figure}

\begin{table}[]
\centering
\caption{Comparison with different denoising method for AlexNet Classification. 
The performance are multi-crop evaluation on the same 1000 images used for denoising experiments.}
\label{classalex}
\resizebox{\columnwidth}{!}
{
\begin{tabular}{llllll}
\hline
\multicolumn{1}{c}{\multirow{2}{*}{Methods}} & \multicolumn{5}{c}{$\sigma$}                                               \\ \cline{2-6} 
\multicolumn{1}{c}{}                         & \multicolumn{1}{c}{15} & \multicolumn{1}{c}{25} & \multicolumn{1}{c}{50} & \multicolumn{1}{c}{90} & \multicolumn{1}{c}{130}\\ \hline
noise free                                   & 59.2 \%                & 59.2 \%                & 59.2 \%                & 59.2 \%                & 59.2 \%                \\
noisy                                        & 57.8 \%                & 50.8 \%                & 30.8 \%                & 9.4  \%                & 2.2  \%                \\
TRND                                         & 56.7 \%                & 53.1 \%                & 46.1 \%                & -                      & -                      \\
CBM3D                                        & 57.9 \%                & 57.7 \%                & 52.2 \%                & 45.4 \%                & 23.8 \%                \\
3DR+VGG-16                                   & 58.5 \%                & 58.1 \%                & 53.7 \%                & 45.7 \%                & 36.1 \%                \\
3DR+AlexNet                                  & 58.5 \%                & 58.1 \%                & 53.6 \%                & 46.4 \%                & 36.3 \%                \\ \hline
\end{tabular}
}
\end{table}

\begin{table}[]
\centering
\caption{Comparison with different denoising method for VGG-16 Classification.
The performance are multi-crop evaluation on the same 1000 images used for denoising experiments.}
\label{classvgg}
\resizebox{\columnwidth}{!}
{
\begin{tabular}{llllll}
\hline
\multicolumn{1}{c}{\multirow{2}{*}{Methods}} & \multicolumn{5}{c}{$\sigma$}                                             \\ \cline{2-6} 
\multicolumn{1}{c}{}                         & \multicolumn{1}{c}{15} & \multicolumn{1}{c}{25} & \multicolumn{1}{c}{50} & \multicolumn{1}{c}{90} & \multicolumn{1}{c}{130}\\ \hline
noise free                                   & 69.6 \%                & 69.6 \%                & 69.6 \%                & 69.6 \%                & 69.6 \%                \\
noisy                                        & 65.8 \%                & 61.3 \%                & 49.6 \%                & 25.4 \%                & 9.1  \%                \\
TRND                                         & 67.1 \%                & 63.3 \%                & 53.2 \%                & -                      & -                      \\
CBM3D                                        & 69.0 \%                & 66.7 \%                & 60.7 \%                & 52.0 \%                & 23.8 \%                \\
3DR+VGG-16                                   & 69.7 \%                & 67.4 \%                & 60.3 \%                & 45.2 \%                & 30.6 \%                \\
3DR+AlexNet                                  & 69.3 \%                & 68.2 \%                & 60.3 \%                & 45.8 \%                & 34.0 \%                \\ \hline
\end{tabular}
}
\end{table}

\begin{table}[]
\centering
\caption{Comparison with 3DR unsupervised denoising for AlexNet and VGG-16.
The performance are multi-crop evaluation on the same 1000 images used for denoising experiments.}
\label{unsupervise}
{
\begin{tabular}{llll}
\hline
\multicolumn{1}{c}{\multirow{2}{*}{Classification}} & \multicolumn{3}{c}{$\sigma$}                                             \\ \cline{2-4} 
\multicolumn{1}{c}{}                                & \multicolumn{1}{c}{15} & \multicolumn{1}{c}{25} & \multicolumn{1}{c}{50} \\ \hline
AlexNet                                             & 69.6 \%                & 69.6 \%                & 11.4 \%                \\
VGG-16                                              & 65.8 \%                & 61.3 \%                & 31.8 \%                \\ \hline
\end{tabular}
}
\end{table}

\subsection{Image classification}
In this section, we study how denoising methods affect classification performance.
We firstly use various denoising methods to obtain clean image then classify the image by AlexNet and VGG-16. 
The results are reported on Tab.~\ref{classalex} and~\ref{classvgg}.
We conclude that denoising methods indeed help improving the classification accuracy.
For example, in the case of $\sigma = 25$, our proposed method 3DR+AlexNet gains 7.3\% and 6.9\% advantage over the case of classification without applying denoising for both AlexNet and VGG-16.
In the case of $\sigma = 50$, we outperform noisy image classification by 22.4\% and 10.7\% respectively.
Tab.~\ref{classalex} and~\ref{classvgg} confirm our intuition that denoising step is indeed useful for classification. 
More importantly, if we can develop more powerful denoising method,
it is more likely to improve the classification method by larger margin. 

On the other hand, what if we let the classification model denoise the corrupted image? \ie we use the softmax layer for classification as our loss function, at the meantime get rid of the denoising loss layer and freeze the weights of classification model.
As it turns out (Tab.~\ref{unsupervise}), denoising by classification performs way worse than applying common denoising method. 
For instance, AlexNet merely achieves 11.4\% accuracy on classification task for $\sigma = 50$.
The performance of VGG-16 also presents huge setback.
Hence, it may not be a good idea to let classification model to make the image clean to itself, denoising model supervised by regression is still necessary. 

\begin{figure}
\centering
\setlength{\tabcolsep}{1pt}
{
\begin{tabular}{ccc}
noisy & TRND & CBM3D \\  
\includegraphics[width=0.32\linewidth]{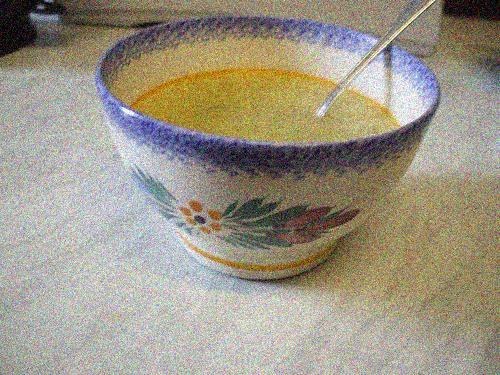}&
\includegraphics[width=0.32\linewidth]{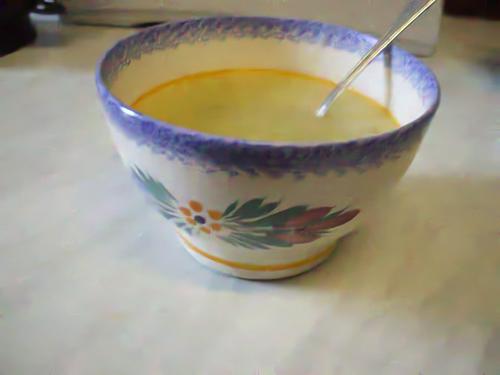}&
\includegraphics[width=0.32\linewidth]{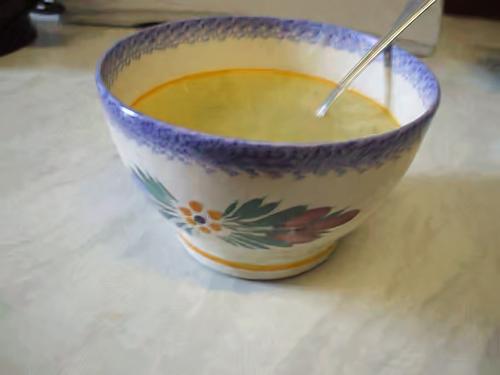}\\
\textbf{0.1922/soup bowl} & 0.2197/cup & 0.2046/eggnog\\

3DR + VGG-16 &3DR + AlexNet & noise free\\
\includegraphics[width=0.32\linewidth]{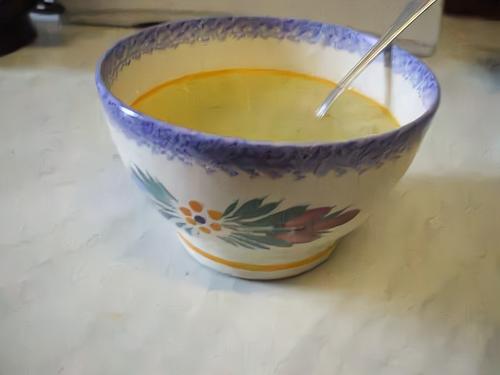}&
\includegraphics[width=0.32\linewidth]{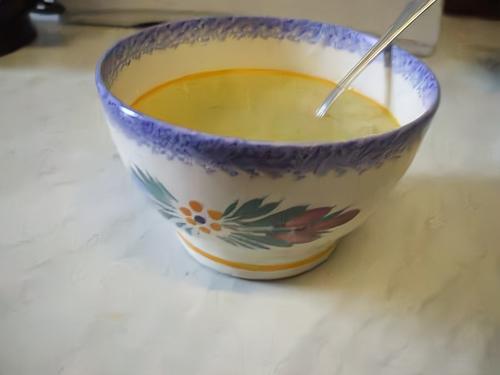}&
\includegraphics[width=0.32\linewidth]{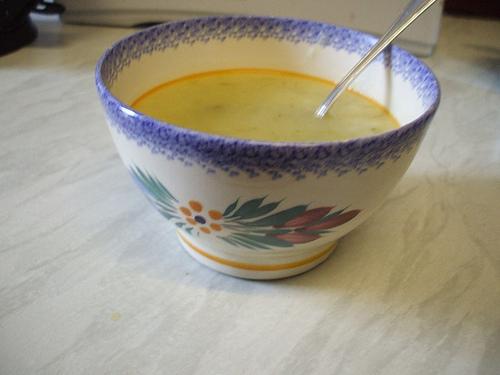}\\
0.2092/cup &0.2116/cup & 0.2079/cup\\
\end{tabular}
}       
\caption{AlexNet example for soup bowl, $\sigma=25$.}
\label{fig:bowl}
\vspace{-0.2cm}
\end{figure}

\begin{figure}
\centering
\setlength{\tabcolsep}{1pt}
{
\begin{tabular}{ccc}
noisy & TRND & CBM3D \\ 
\includegraphics[width=0.32\linewidth]{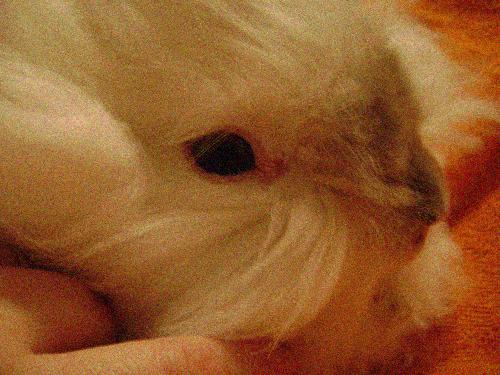}&
\includegraphics[width=0.32\linewidth]{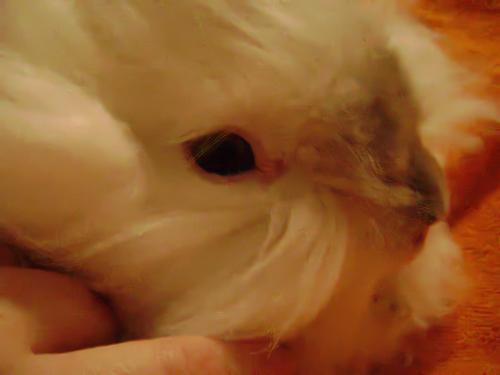}&
\includegraphics[width=0.32\linewidth]{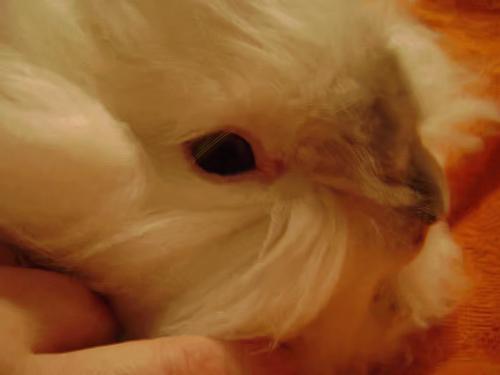}\\
\textbf{0.3006/Angora} & 0.1340/Hamster & 0.3074/Maltese\\

3DR + VGG-16 &3DR + AlexNet & noise free\\
\includegraphics[width=0.32\linewidth]{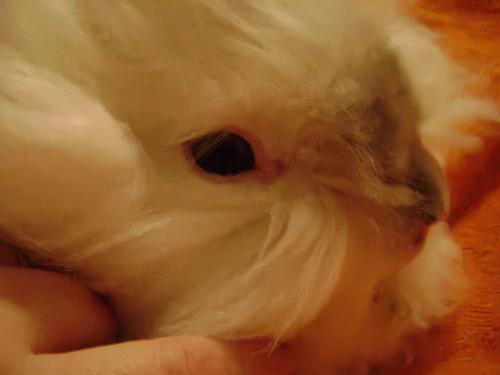}&
\includegraphics[width=0.32\linewidth]{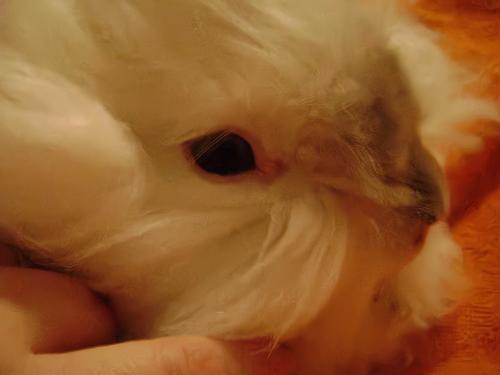}&
\includegraphics[width=0.32\linewidth]{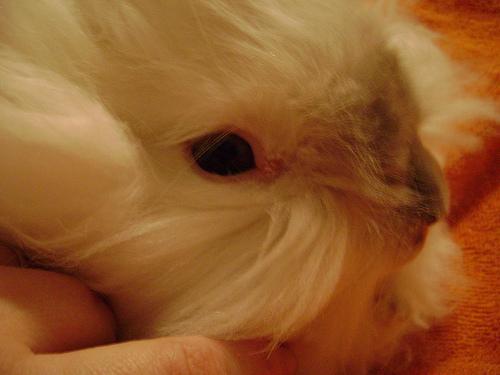}\\
0.2112/Maltese &0.2940/Maltese & 0.4564/Maltese\\
\end{tabular}
}       
\caption{VGG-16 example for Angora, $\sigma=15$.}
\label{fig:rabbit}
\vspace{-0.2cm}
\end{figure}

\textbf{Sometimes Noise helps.}
Besides, we find out that for certain cases AlexNet/VGG-16 fail to recognize neither the noise free nor denoised image.
However, they can classify noisy images correctly. 
For instance, for AlexNet there exists 2.5\% 3.5\% 3.1\% images under the condition $\sigma = 15, 25, 30$, for VGG-16 the rate is 1.4\%, 1.8\%, 1.7\% respectively.
Given there are 1000 categories for ImageNet, such rate of noise-improve-classify phenomenon cannot be simply considered as coincidence, 
on the contrary, it shows that AlexNet and VGG-16 are highly non-linear, 
quite sensitive to the noise.
As shown in Fig.~\ref{fig:bowl}, AlexNet recognize the noisy soup bowl image with 19.22\% confidence rate,
but it cannot classify other images correctly.
Fig.~\ref{fig:rabbit} demonstrate a similar example as well, 
noisy image is accurately classified to be Angora for VGG-16 with 30.06\% confidence rate.
Such counter-intuitive examples deserves more thorough study in later work,
such that we can boost and stable neural network model.

\section{Conclusion}
\label{sec:conclusion}
In this paper, we proposed a novel color image denoising CNN model composed from a novel 3D residual learning stage and a standard top classification architecture (AlexNet/VGG-16). Experimental results on large and diverse datasets show that our model outperforms state-of-the-art methods.
Meanwhile, we also study how denoising methods affect the classification models and conclude that denoising can indeed recover the classification accuracy. Last but not least, we notice that in certain cases noise can actually help classification models to `see' the image clearer, 
which motivates us to study how to robustify neural networks by various noises in future work. 

\bibliographystyle{ieee}
\bibliography{egbib}

\begin{thebibliography}{10}\itemsep=-1pt

\bibitem{brezis2010functional}
H.~Brezis.
\newblock {\em Functional analysis, Sobolev spaces and partial differential
  equations}.
\newblock Springer Science \& Business Media, 2010.

\bibitem{Burger-CVPR-2012}
H.~Burger, C.~Schuler, and S.~Harmeling.
\newblock Image denoising: Can plain neural networks compete with bm3d?
\newblock In {\em IEEE Computer Vision and Pattern Recognition}, pages
  2392--2399, 2012.

\bibitem{chen2013revisiting}
Y.~Chen, T.~Pock, R.~Ranftl, and H.~Bischof.
\newblock Revisiting loss-specific training of filter-based mrfs for image
  restoration.
\newblock In {\em Pattern Recognition}, pages 271--281. Springer, 2013.

\bibitem{chen2015learning}
Y.~Chen, W.~Yu, and T.~Pock.
\newblock On learning optimized reaction diffusion processes for effective
  image restoration.
\newblock In {\em Proceedings of the IEEE Conference on Computer Vision and
  Pattern Recognition}, pages 5261--5269, 2015.

\bibitem{dabov2007color}
K.~Dabov, A.~Foi, V.~Katkovnik, and K.~Egiazarian.
\newblock Color image denoising via sparse 3d collaborative filtering with
  grouping constraint in luminance-chrominance space.
\newblock In {\em 2007 IEEE International Conference on Image Processing},
  volume~1, pages I--313. IEEE, 2007.

\bibitem{Dabov-TIP-2007}
K.~Dabov, A.~Foi, V.~Katkovnik, and K.~Egiazarian.
\newblock Image denoising by sparse 3d transform-domain collaborative
  filtering.
\newblock {\em IEEE Trans. Image Processing}, 16:2080--2095, 2007.

\bibitem{deng2009imagenet}
J.~Deng, W.~Dong, R.~Socher, L.-J. Li, K.~Li, and L.~Fei-Fei.
\newblock Imagenet: A large-scale hierarchical image database.
\newblock In {\em Computer Vision and Pattern Recognition, 2009. CVPR 2009.
  IEEE Conference on}, pages 248--255. IEEE, 2009.

\bibitem{everingham2015pascal}
M.~Everingham, S.~A. Eslami, L.~Van~Gool, C.~K. Williams, J.~Winn, and
  A.~Zisserman.
\newblock The pascal visual object classes challenge: A retrospective.
\newblock {\em International Journal of Computer Vision}, 111(1):98--136, 2015.

\bibitem{Gu-CVPR-2014}
S.~Gu, L.~Zhang, W.~Zuo, and X.~Feng.
\newblock Weighted nuclear norm minimization with application to image
  denoising.
\newblock In {\em CVPR}, 2014.

\bibitem{he2015deep}
K.~He, X.~Zhang, S.~Ren, and J.~Sun.
\newblock Deep residual learning for image recognition.
\newblock {\em arXiv preprint arXiv:1512.03385}, 2015.

\bibitem{ji20133d}
S.~Ji, W.~Xu, M.~Yang, and K.~Yu.
\newblock 3d convolutional neural networks for human action recognition.
\newblock {\em Pattern Analysis and Machine Intelligence, IEEE Transactions
  on}, 35(1):221--231, 2013.

\bibitem{kim2015accurate}
J.~Kim, J.~K. Lee, and K.~M. Lee.
\newblock Accurate image super-resolution using very deep convolutional
  networks.
\newblock {\em arXiv preprint arXiv:1511.04587}, 2015.

\bibitem{kingma2014adam}
D.~Kingma and J.~Ba.
\newblock Adam: A method for stochastic optimization.
\newblock {\em arXiv preprint arXiv:1412.6980}, 2014.

\bibitem{krizhevsky2012imagenet}
A.~Krizhevsky, I.~Sutskever, and G.~E. Hinton.
\newblock Imagenet classification with deep convolutional neural networks.
\newblock In {\em Advances in neural information processing systems}, pages
  1097--1105, 2012.

\bibitem{lecun1989backpropagation}
Y.~LeCun, B.~Boser, J.~S. Denker, D.~Henderson, R.~E. Howard, W.~Hubbard, and
  L.~D. Jackel.
\newblock Backpropagation applied to handwritten zip code recognition.
\newblock {\em Neural computation}, 1(4):541--551, 1989.

\bibitem{lin2014microsoft}
T.-Y. Lin, M.~Maire, S.~Belongie, J.~Hays, P.~Perona, D.~Ramanan,
  P.~Doll{\'a}r, and C.~L. Zitnick.
\newblock Microsoft coco: Common objects in context.
\newblock In {\em European Conference on Computer Vision}, pages 740--755.
  Springer, 2014.

\bibitem{Mairal-ICCV-2009}
J.~Mairal, F.~Bach, J.~Ponce, G.~Sapiro, and A.~Zisserman.
\newblock Non-local sparse models for image restoration.
\newblock In {\em IEEE 12th International Conference on Computer Vision}, pages
  2272--2279, 2009.

\bibitem{martin2001database}
D.~Martin, C.~Fowlkes, D.~Tal, and J.~Malik.
\newblock A database of human segmented natural images and its application to
  evaluating segmentation algorithms and measuring ecological statistics.
\newblock In {\em Computer Vision, 2001. ICCV 2001. Proceedings. Eighth IEEE
  International Conference on}, volume~2, pages 416--423. IEEE, 2001.

\bibitem{nguyen2015deep}
A.~Nguyen, J.~Yosinski, and J.~Clune.
\newblock Deep neural networks are easily fooled: High confidence predictions
  for unrecognizable images.
\newblock In {\em 2015 IEEE Conference on Computer Vision and Pattern
  Recognition (CVPR)}, pages 427--436. IEEE, 2015.

\bibitem{rajwade2013image}
A.~Rajwade, A.~Rangarajan, and A.~Banerjee.
\newblock Image denoising using the higher order singular value decomposition.
\newblock {\em IEEE Transactions on Pattern Analysis and Machine Intelligence},
  35(4):849--862, 2013.

\bibitem{schmidt2014cascades}
U.~Schmidt, J.~Jancsary, S.~Nowozin, S.~Roth, and C.~Rother.
\newblock Cascades of regression tree fields for image restoration.
\newblock 2014.

\bibitem{schmidt2014shrinkage}
U.~Schmidt and S.~Roth.
\newblock Shrinkage fields for effective image restoration.
\newblock In {\em Proceedings of the IEEE Conference on Computer Vision and
  Pattern Recognition}, pages 2774--2781, 2014.

\bibitem{simonyan2014very}
K.~Simonyan and A.~Zisserman.
\newblock Very deep convolutional networks for large-scale image recognition.
\newblock {\em arXiv preprint arXiv:1409.1556}, 2014.

\bibitem{szegedy2015going}
C.~Szegedy, W.~Liu, Y.~Jia, P.~Sermanet, S.~Reed, D.~Anguelov, D.~Erhan,
  V.~Vanhoucke, and A.~Rabinovich.
\newblock Going deeper with convolutions.
\newblock In {\em Proceedings of the IEEE Conference on Computer Vision and
  Pattern Recognition}, pages 1--9, 2015.

\bibitem{szegedy2013intriguing}
C.~Szegedy, W.~Zaremba, I.~Sutskever, J.~Bruna, D.~Erhan, I.~Goodfellow, and
  R.~Fergus.
\newblock Intriguing properties of neural networks.
\newblock {\em arXiv preprint arXiv:1312.6199}, 2013.

\bibitem{Timofte-CVPR-2016}
R.~Timofte, R.~Rothe, and L.~Van~Gool.
\newblock Seven ways to improve example-based single image super resolution.
\newblock In {\em The IEEE Conference on Computer Vision and Pattern
  Recognition (CVPR)}, June 2016.

\bibitem{timofte2013anchored}
R.~Timofte, V.~Smet, and L.~Gool.
\newblock Anchored neighborhood regression for fast example-based
  super-resolution.
\newblock In {\em Proceedings of the IEEE International Conference on Computer
  Vision}, pages 1920--1927, 2013.

\bibitem{vemulapalli2015deep}
R.~Vemulapalli, O.~Tuzel, and M.-Y. Liu.
\newblock Deep gaussian conditional random field network: A model-based deep
  network for discriminative denoising.
\newblock {\em arXiv preprint arXiv:1511.04067}, 2015.

\bibitem{zagoruyko2016wide}
S.~Zagoruyko and N.~Komodakis.
\newblock Wide residual networks.
\newblock {\em arXiv preprint arXiv:1605.07146}, 2016.

\bibitem{zeyde2010single}
R.~Zeyde, M.~Elad, and M.~Protter.
\newblock On single image scale-up using sparse-representations.
\newblock In {\em Curves and Surfaces}, pages 711--730. Springer, 2010.

\bibitem{zhou2014learning}
B.~Zhou, A.~Lapedriza, J.~Xiao, A.~Torralba, and A.~Oliva.
\newblock Learning deep features for scene recognition using places database.
\newblock In {\em Advances in neural information processing systems}, pages
  487--495, 2014.

\bibitem{Zoran-ICCV-2011}
D.~Zoran and Y.~Weiss.
\newblock From learning models of natural image patches to whole image
  restoration.
\newblock In {\em IEEE International Conference on Computer Vision}, pages
  479--486, 2011.

\end{thebibliography}

\begin{IEEEbiography}[{\includegraphics[width=1in,height=1.25in,clip,keepaspectratio]{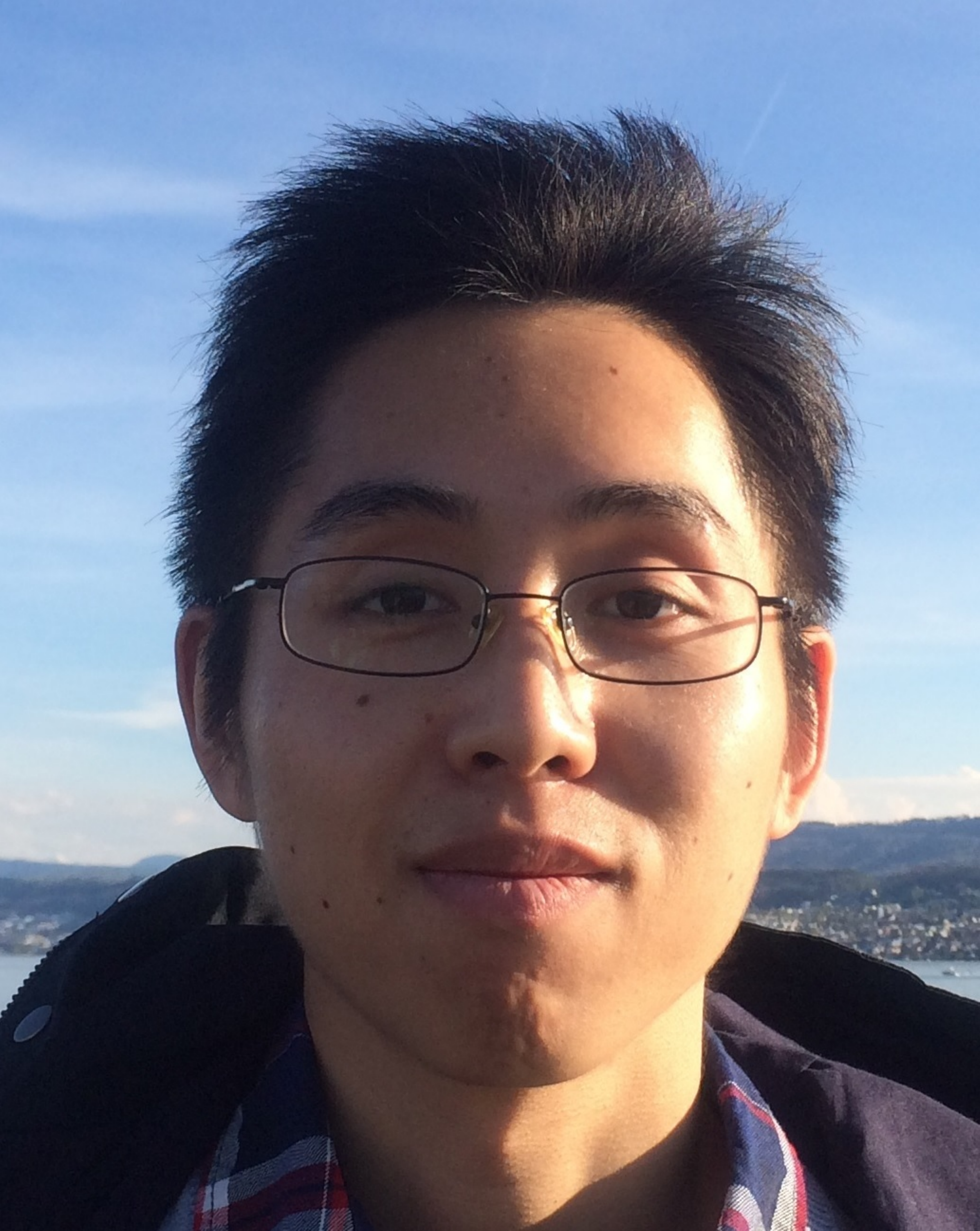}}]{Jiqing Wu} received the B.S. degree in mechanical engineering from Shanghai Maritime University, China, in 2006, the B.S. degree from TU Darmstadt, Germany, in 2012,
the M.Sc. degree from ETH Zurich, Switzerland, in 2015, in mathematics, respectively.
He is currently pursuing a PhD degree under the supervision of prof. Luc Van Gool, in his lab at ETH Zurich. 
His research interests mainly concern image demosaicing, image restoration, and early vision. 
\end{IEEEbiography}

\begin{IEEEbiography}[{\includegraphics[width=1in,height=1.25in,clip,keepaspectratio]{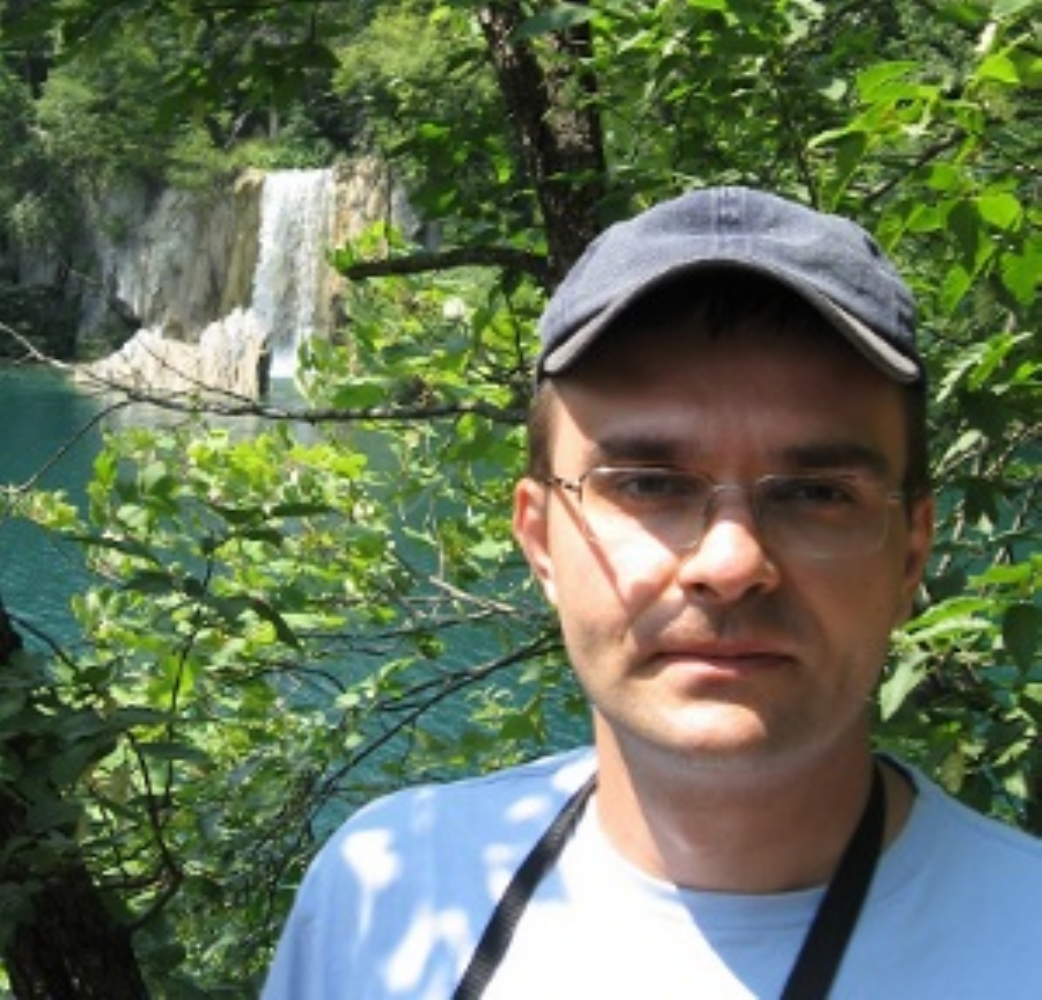}}]{Radu Timofte} obtained the PhD degree in Electrical Engineering at the KU Leuven, Belgium, in 2013, the M.Sc. at the Univ. of Eastern Finland in 2007 and the Dipl.-Ing. at the Technical Univ. of Iasi, Romania, in 2006. He worked as a researcher for the Univ. of Joensuu (2006-2007), the Czech Technical Univ. (2008) and the KU Leuven (2008-2013). Since 2013 he is postdoc at ETH Zurich, Switzerland, in the lab of prof. Luc Van Gool. He serves as a reviewer for major journals such as PAMI, TIP, IJCV, TNNLS, TKDE, PRL, and T-ITS, and conferences such as CVPR, ICCV, ECCV, and NIPS. He received best paper awards at ICPR 2012, at CVVT workshop from ECCV 2012 and at ChaLearn workshop from ICCV 2015. His current research interests include sparse and collaborative representations, image restoration and enhancement, and multi-view object class recognition.
\end{IEEEbiography}


\begin{IEEEbiography}[{\includegraphics[width=1in,height=1.25in,clip,keepaspectratio]{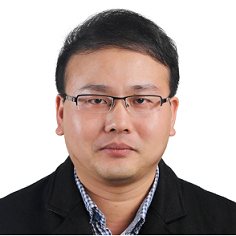}}]{Zhiwu Huang} received the B.S. degree in
computer science and technology from Huaqiao
University, Quanzhou, China, in 2007, the
M.S. degree in computer software and theory from
Xiamen University, Xiamen, China, in 2010, and the
Ph.D. degree in computer science and technology
from the Institute of Computing Technology,
Chinese Academy of Sciences, Beijing, China,
in 2015.
He has been a Post-Doctoral Researcher with
the Computer Vision Laboratory, Swiss Federal
Institute of Technology, Zurich, Switzerland, since 2015. His current research
interests include Riemannian geometry, kernel learning, metric learning and
feature learning with application to face recognition, face retrieval, facial
expression recognition, gesture recognition, action recognition, and image
set classification in computer vision. 
\end{IEEEbiography}

\begin{IEEEbiography}[{\includegraphics[width=1in,height=1.25in,clip,keepaspectratio]{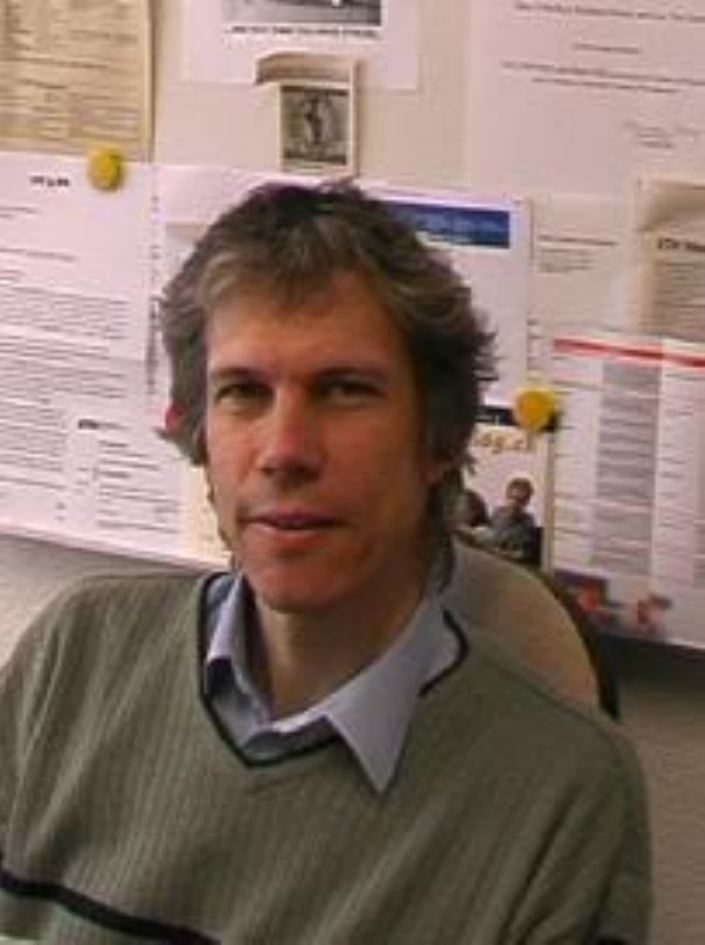}}]{Luc Van Gool} obtained the master and PhD degrees in Electrical Engineering at the KU Leuven, Belgium, resp. in 1982 and 1991. He is a full professor for Computer Vision at both KU Leuven and ETH Zurich. With his two research labs, he focuses on object recognition, tracking and gesture analysis, and 3D acquisition and modeling. Luc Van Gool was a program chair of ICCV 2005, and general chair of ICCV 2011, and acted as GC of ECCV 2014. He is an editor-in-chief of Foundations and Trends in Computer Graphics and Vision. He also is a co-founder of the spin-off companies Eyetronics, GeoAutomation, kooaba, procedural, eSaturnus, upicto, Fashwell, Merantix, Spectando, and Parquery. He received several best paper awards, incl. at ICCV 1998, CVPR 2007, ACCV 2007, ICRA 2009, BMVC 2011, and ICPR 2012.
\end{IEEEbiography}




\end{document}